\documentclass[journal]{IEEEtran}

\usepackage{cite}
\usepackage{graphicx}
\usepackage{multirow}
\usepackage{amssymb}
\usepackage[pagebackref=true,breaklinks=true,letterpaper=true,colorlinks,bookmarks=false]{hyperref}

\begin{document}

\title{PEPSI++: Fast and Lightweight Network \\ for Image Inpainting 
}




\author{Yong-Goo~Shin, Min-Cheol~Sagong, Yoon-Jae~Yeo, Seung-Wook~Kim, and Sung-Jea Ko,~\IEEEmembership{Fellow}, IEEE
\thanks{Y.-G. Shin is with the School of Electrical Engineering Department, Korea University, Anam-dong, Sungbuk-gu, Seoul, 136-713, Rep. of Korea (e-mail: ygshin@dali.korea.ac.kr).}
\thanks{M.-C. Sagong is with the School of Electrical Engineering Department, Korea University, Anam-dong, Sungbuk-gu, Seoul, 136-713, Rep. of Korea (e-mail: mcsagong@dali.korea.ac.kr).}
\thanks{Y.-J. Yeo is with the School of Electrical Engineering Department, Korea University, Anam-dong, Sungbuk-gu, Seoul, 136-713, Rep. of Korea (e-mail: yjyeo@dali.korea.ac.kr).}
\thanks{S.-W. Kim is with the School of Electrical Engineering Department, Korea University, Anam-dong, Sungbuk-gu, Seoul, 136-713, Rep. of Korea (e-mail: swkim@dali.korea.ac.kr).}
\thanks{S.-J. Ko is with the School of Electrical Engineering Department, Korea University, Anam-dong, Sungbuk-gu, Seoul, 136-713, Rep. of Korea (e-mail: sjko@korea.ac.kr).}}


\markboth{Submitted to IEEE transactions on Neural Networks and Learning System}%
{Shell \MakeLowercase{\textit{Shin et al.}}}

\maketitle

\begin{abstract}
Among the various generative adversarial network (GAN)-based image inpainting methods, coarse-to-fine network with a contextual attention module (CAM) has shown remarkable performance. However, owing to two stacked generative networks, the coarse-to-fine network needs numerous computational resources such as convolution operations and network parameters, which result in low speed. To address this problem, we propose a novel network architecture called PEPSI: parallel extended-decoder path for semantic inpainting network, which aims at reducing the hardware costs and improving the inpainting performance. PEPSI consists of a single shared encoding network and parallel decoding networks called coarse and inpainting paths. The coarse path produces a preliminary inpainting result to train the encoding network for the prediction of features for the CAM. Simultaneously, the inpainting path generates higher inpainting quality using the refined features reconstructed via the CAM. In addition, we propose Diet-PEPSI that significantly reduces the network parameters while maintaining the performance. In Diet-PEPSI, to capture the global contextual information with low hardware costs, we propose novel rate-adaptive dilated convolutional layers, which employ the common weights but produce dynamic features depending on the given dilation rates. Extensive experiments comparing the performance with state-of-the-art image inpainting methods demonstrate that both PEPSI and Diet-PEPSI improve the qualitative scores, \textit{i.e.} the peak signal-to-noise ratio (PSNR) and structural similarity (SSIM), as well as significantly reduce hardware costs such as computational time and the number of network parameters. 


\end{abstract}

\section{Introduction}
\label{sec1}
\IEEEPARstart{I}{mage} inpainting techniques which attempt to remove an unwanted object or synthesize missing parts of an image have attracted wide-spread interest in computer vision and graphics communities~\cite{bertalmio2000image, efros2001image, barnes2009patchmatch, yu2018generative, noori2010convolution, li2017context, iizuka2017globally, kohler2014mask, li2016combining, pathak2016context, yang2017high, fawzi2016image, cai2017blind, yeh2017semantic, liu2018image, song2018contextual, wang2018image}. Recent studies used the generative adversarial network (GAN) to produce appropriate structures for the missing regions, $i.e.$ hole regions~\cite{kohler2014mask, li2016combining, xu2014deep}. Among the recent state-of-the-art inpainting methods, the coarse-to-fine network with a contextual attention module (CAM) has shown remarkable performance~\cite{yu2018generative, yu2018free}. This network is composed of two stacked generative networks including the coarse network and refinement one. The coarse network roughly fills the hole regions using a simple dilated convolutional network trained with reconstruction loss. The refinement network improves the quality of the roughly completed image by using the CAM that generates feature patches of the hole regions by borrowing information from distant spatial locations. Despite the promising results, the coarse-to-fine network requires high computational resources and consumes considerable memories.

In previous work~\cite{PEPSI}, we introduced a novel network structure called PEPSI: parallel extended-decoder path for semantic inpainting, which aims at reducing the number of convolution operations as well as improving the inpainting performance. PEPSI is composed of a single encoding network and parallel decoding networks which consist of coarse and inpainting paths. The coarse path generates a preliminary inpainting result to train the encoding network for prediction of features for the CAM. At the same time, the inpainting path produces image with high quality using the refined features reconstructed via the CAM. To make a single encoding network handle two different tasks, which are feature extraction for both roughly completed and high-quality results, we propose a joint learning technique that jointly optimizes two different paths. This learning scheme facilitates the generation of high-quality inpainting image without the stacked generative networks, $i.e.$ the coarse-to-fine network.

\begin{figure*}[!ht]
\centering
\includegraphics[width=0.95\textwidth]{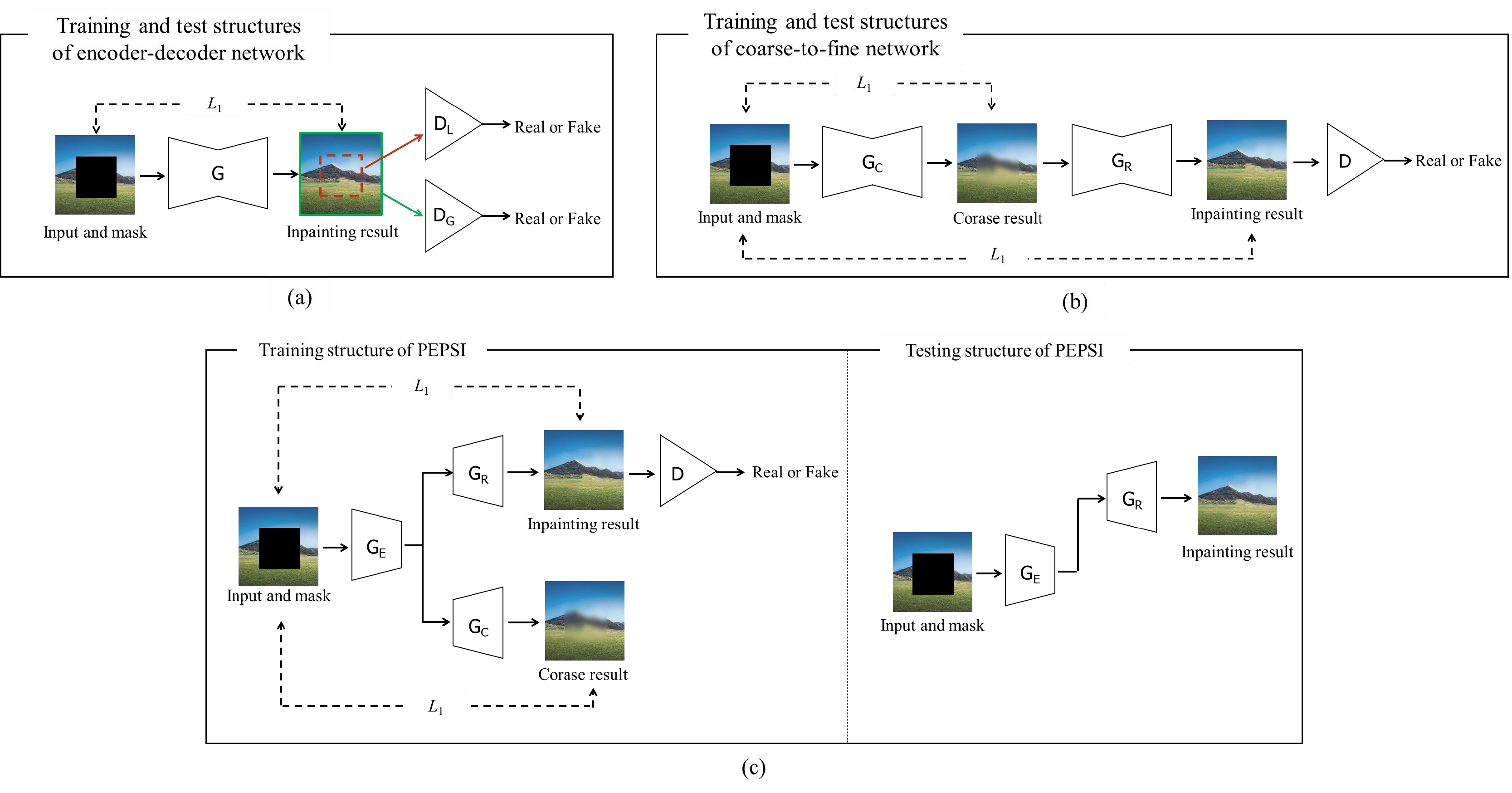}
\caption{Overview of the network architectures of the conventional and proposed methods, where D and G indicate the discriminator and generator, respectively. (a) Architecture of traditional encoder-decoder network~\cite{iizuka2017globally}. (b) Architecture of coarse-to-fine network~\cite{yu2018generative, yu2018free}. (c) Architecture of PEPSI.}
\label{fig:fig0}
\vspace{-0.3cm}
\end{figure*}

Although PEPSI exhibits faster operation speed compared with conventional methods, it still needs substantial memory owing to a series of dilated convolutional layers in the encoding network, which occupies nearly 67 percent of network parameters. The intuitive way to save memory consumption is to prune channels in the dilated convolutional layers; however, it often results in inferior results. To address this challenge, this paper presents an extended version of PEPSI, called Diet-PEPSI, which significantly reduces the network parameters while retaining the inpainting performance. In Diet-PEPSI, we propose novel rate-adaptive dilated convolutional layers which require low hardware costs by sharing the weights in every layer but generate dynamic features according to the given dilation rates. More specifically, to produce the rate-specific features, the rate-adaptive dilated convolutional layers modulate the shared weights by differently scaling and shifting according to the given dilation rates. Since the rate-adaptive dilated convolutional layers share the weights with each other, the number of network parameters can be significantly reduced compared with multiple standard dilated convolutional layers. In this paper, we apply the proposed rate-adaptive dilated convolutional layers to Diet-PEPSI using residual blocks~\cite{he2016deep} called Diet-PEPSI units (DPUs). By replacing the multiple dilated convolutional layers with DPUs, Diet-PEPSI covers the same size of the receptive field with a smaller number of parameters than PEPSI.

Furthermore, we investigate an obstacle with the discriminator in traditional GAN-based image inpainting methods~\cite{goodfellow2014generative, yeh2017semantic}. In general, conventional methods employ global and local discriminators trained with a combined loss, the $L_2$ pixel-wise reconstruction loss and adversarial loss, which assists the networks in generating a more natural image by minimizing the difference between the reference and the inpainted images. More specifically, the global discriminator takes the whole image as input to recognize global consistency, whereas the local one only views at a small region around the hole in order to judge the quality of more detailed appearance. However, the local discriminator has a drawback that it can only deal with a single rectangular hole region. In other words, since the holes can appear with arbitrary shapes, sizes, and locations in real-world applications, the local discriminator is difficult to apply to the inpainting network for inpainting the holes with irregular shapes. To solve this problem, we propose a region ensemble discriminator (RED) which integrates the global and local discriminators. Since each pixel in the last layer has a different receptive field in the image domain, the RED adopts individual fully connected layers on each pixel in the last convolutional layer. By individually computing an adversarial loss in each pixel, the RED can deal with the various holes with arbitrary shapes.

In summary, this paper has three major contributions. (\textit{i}) We propose a novel network architecture called PEPSI that achieves superior performance compared to conventional methods as well as significantly reduces the operation time. (\textit{ii}) We propose Diet-PEPSI that applies novel rate-adaptive convolution layers to further reduce the hardware costs while maintaining the overall quality of the results, which makes the proposed method compatible with the hardware. (\textit{iii}) A novel discriminator, called RED, is proposed to handle both squared and irregular hole regions for real applications. In the remainder of this paper, we introduce the related work and preliminaries in Section~\ref{sec2} and Section~\ref{sec3}, respectively. The PEPSI and Diet-PEPSI are discussed in Section~\ref{sec4}. In Section~\ref{sec5}, extensive experimental results are presented to demonstrate that the proposed method outperforms conventional methods on various datasets such as CelebA~\cite{karras2017progressive, liu2015deep}, Place2~\cite{zhou2018places}, and ImageNet~\cite{krizhevsky2012imagenet}. Finally, the conclusion is provided in Section~\ref{sec6}.


\section{Related work}
\label{sec2}

Existing image inpainting techniques can be divided into two groups~\cite{yu2018generative}: traditional and deep learning-based methods. The traditional techniques include diffusion-based and patch-based methods. The diffusion-based method fills the hole regions by propagating the local image appearance around the holes~\cite{bertalmio2000image, efros2001image, yu2018generative, noori2010convolution}. The diffusion-based method performs well on the small and narrow holes, but often fails to fill complex hole regions such as faces and objects with non-repetitive structures. In contrast, the patch-based technique results in the better performance in filling the complicated images with large hole regions~\cite{yu2018generative, simakov2008summarizing}. This method samples texture patches from the existing regions of image, \textit{i.e.} background regions, and pastes them into the hole region. Barnes \textit{et al.}~\cite{barnes2009patchmatch} introduced a fast approximate nearest neighbor patch search algorithm, called PatchMatch, which exhibited notable performance for image editing applications such as the image inpainting. However, PatchMatch often fills the hole regions regardless of the visual semantics or the global structure of an image, which results in the resultant images with poor visual quality. 

\begin{figure}[t]
\centering
\includegraphics[width=1.0\linewidth]{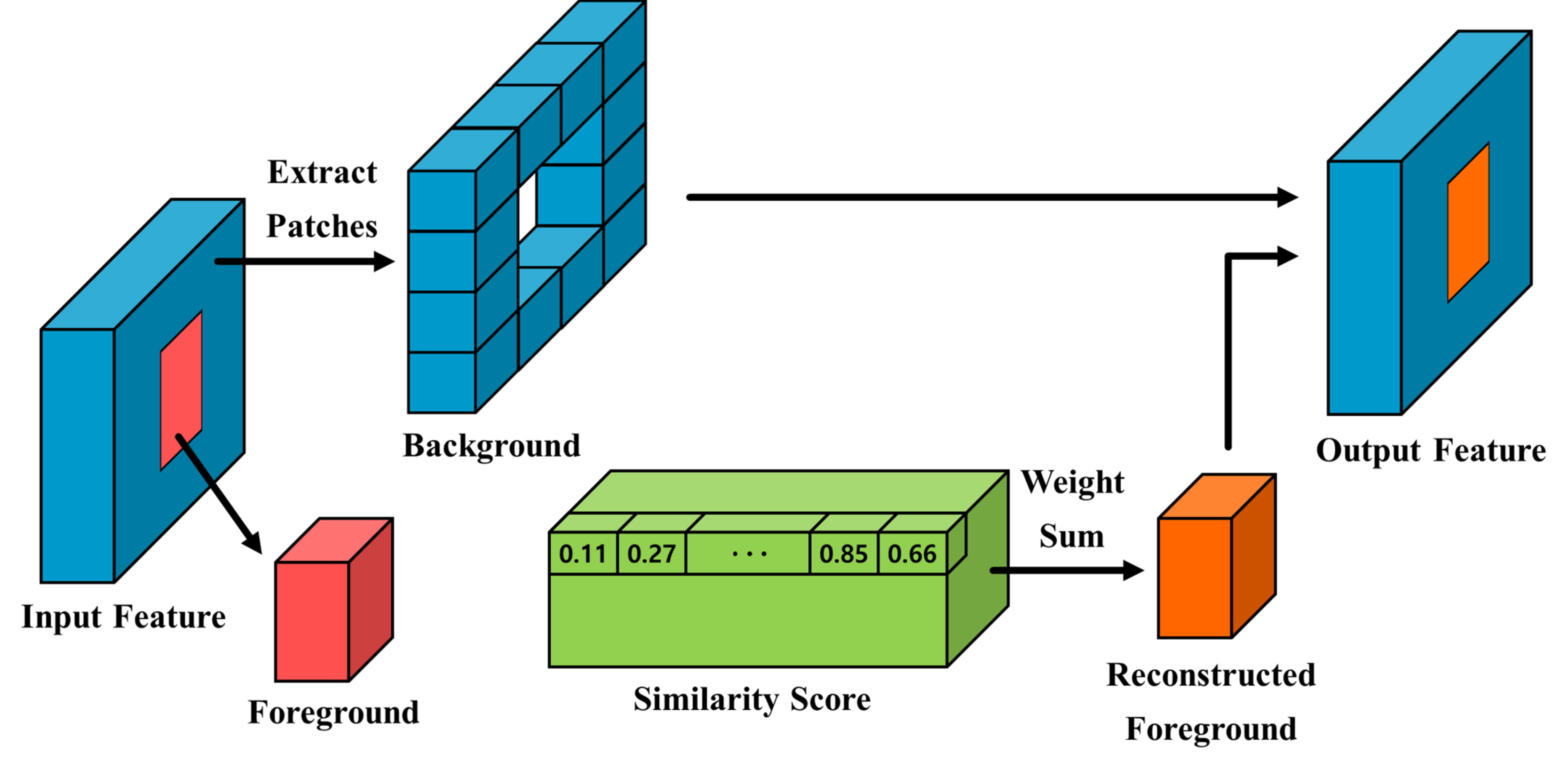}
\caption{Illustration of the CAM. The conventional CAM reconstructs foreground patches by measuring the cosine similarities with background patches. In contrast, the modified CAM uses the Euclidean distance to compute similarity scores.}
\label{fig:fig1}
\vspace{-0.3cm}
\end{figure}

By using the convolutional neural network (CNN), the deep learning-based method learns how to extract semantic information for producing the structures of the hole regions~\cite{kohler2014mask, li2016combining, xu2014deep}. The CNN-based image inpainting methods employing an encoder-decoder structure have shown superior performance on inpainting the complex hole region compared with the diffusion- or patch-based methods~\cite{kohler2014mask, xu2014deep}. However, these methods often generate an image with visual artifacts such as boundary artifacts and blurry texture inconsistent with surrounding areas. To alleviate this problem, Pathak \textit{et al.}~\cite{pathak2016context} adopted the GAN~\cite{goodfellow2014generative} to enhance the coherence between the background and hole regions. They trained the entire network using a combined loss, the $L_2$ pixel-wise reconstruction loss and adversarial loss, which drives the networks to minimize the difference between the reference and inpainted images as well as to produce plausible new contents in highly structured images such as faces and scenes. However, this method has a limitation that it only can fill the square hole located at the center of an image.

To inpaint the images with square hole in arbitrary locations, as shown in Fig.~\ref{fig:fig0}(a), Iizuka \textit{et al.}~\cite{iizuka2017globally} proposed an improved network structure which employs two sibling discriminators: global and local discriminators. More specifically, the local discriminator only considers the inpainted region to classify the local texture consistency, whereas the global discriminator inspects that the resultant image is consistent across the whole image. Recently, Yu \textit{et al.}~\cite{yu2018generative} have extended this work by using the coarse-to-fine network and the CAM. In particular, by computing the cosine similarity between the background and foreground feature patches, the CAM learns where to borrow the background features for the hole region. In order to collect the background features involved with the missing region, the CAM requires the features at the missing region encoded from roughly completed images. Thus, as shown in Fig.~\ref{fig:fig0}(b), this method employs two stacked generative networks (coarse and refinement networks) to generate an intermediate result, \textit{i.e.} the coarse result, and an inpainting result refined through the refinement network having the CAM. This method achieved remarkable performance compared with the recent state-of-the-art inpainting methods; however, it requires considerable computational resources owing to the two stacked generative networks.



\begin{figure}[t]
\centering
\includegraphics[width=0.9\linewidth]{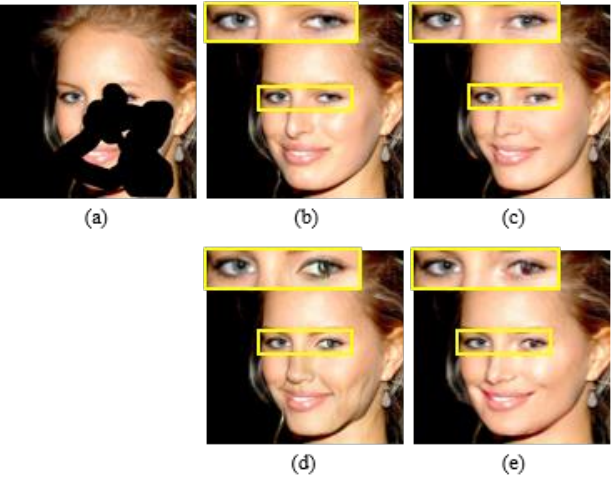}
\caption{Toy examples for the coarse network. (a) Masked input image. (b) Original image. (c) Result from the coarse-to-fine network. (d) Result without the coarse result. (e) Result with LR coarse path.}
\label{fig: fig2}
\vspace{-0.0cm}
\end{figure}

\begin{table}[t]
\caption{Experimental results with GatedConv (GC)~\cite{yu2018free}  using different coarse path. GC* indicates a model trained without coarse results and GC{\dag} indicates a model trained with simplified coarse path.}
\footnotesize
\begin{center}
\begin{tabular}{cccccc}
\hline
 & \multicolumn{2}{c}{Square mask} & \multicolumn{2}{c}{Free-form mask} & \multirow{2}*{Time} \\ \cline{2-5}
  & PSNR & SSIM & PSNR & SSIM & \\
\hline\hline
GC &24.67 &0.8949 &27.78 &0.9252 & 21.39 ms\\
GC$^*$ &23.50 &0.8822 &26.35 &0.9098 & 14.28 ms\\
GC$^{\dag}$ &23.71 &0.8752 &26.22 &0.9026 & 13.32 ms\\
\hline
\end{tabular}
\end{center}
\label{table:table1}
\vspace{-0.3cm}
\end{table}

\begin{figure*}[!ht]
\centering
\includegraphics[width=0.8\textwidth]{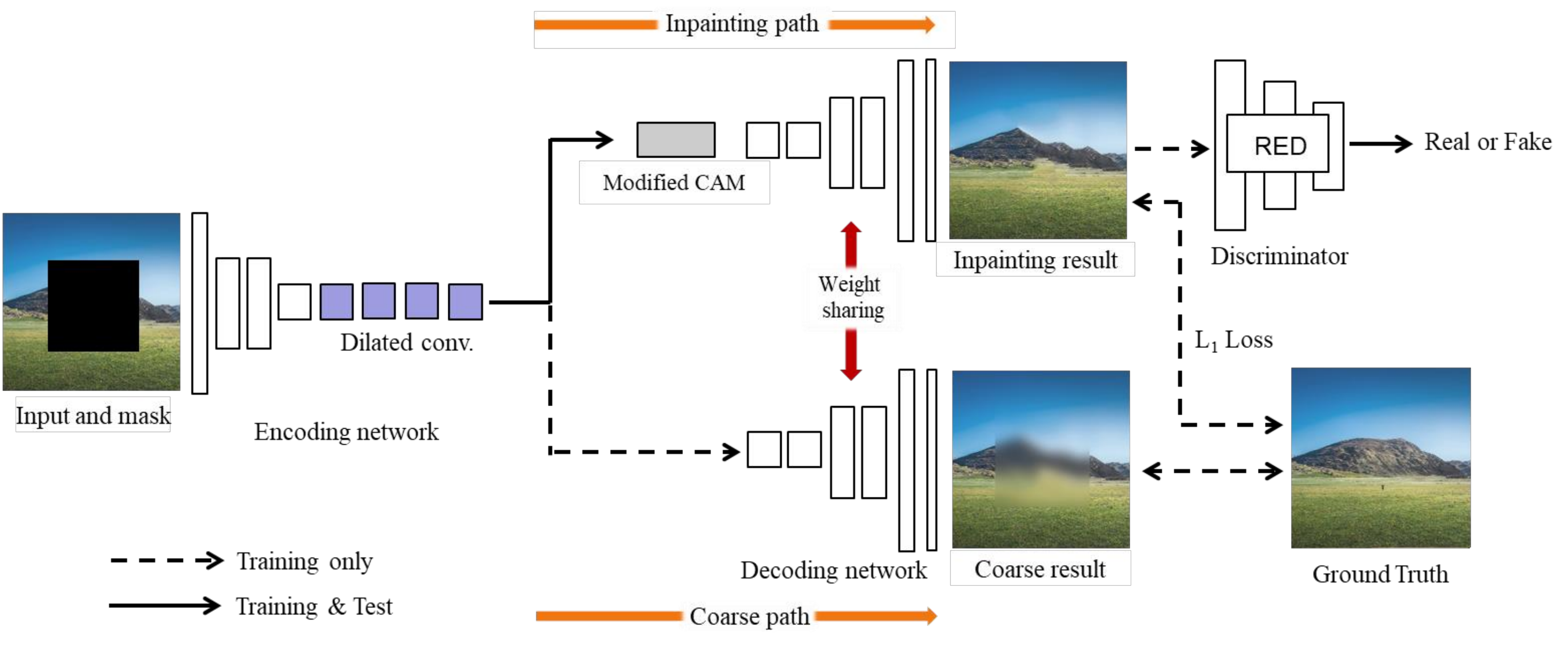}
\caption{An architecture of PEPSI. The coarse path and inpainting path share their weights to improve each other. The coarse path is trained only with the $\textit{L}_1$ reconstruction loss while the inpainting path is trained with both of $\textit{L}_1$ and adversarial losses.}
\label{fig:fig3}
\vspace{-0.3cm}
\end{figure*}

\section{Preliminaries}
\label{sec3}
\subsection{Generative adversarial networks}
\label{sec3.1}
The GAN was first introduced by Goodfellow ~\textit{et al.}~\cite{goodfellow2014generative} for image generation. In general, GAN consists of a generator \textit{G} and a discriminator \textit{D} which are trained with competing goals. The generator is trained to produce a new image, indistinguishable from real images, while the discriminator is optimized to differentiate between real and generated images. Formally, the \textit{G} (\textit{D}) tries to minimize (maximize) the loss function, $i.e.$ adversarial loss, as follows: 

\begin{eqnarray}
\label{eq:gan}
    \lefteqn{\min_G \max_D E_{x\sim P_\mathrm{data}(x)}[\log D(x)]}\nonumber\\
    & & \qquad\qquad\qquad{} +E_{z\sim P_{z(z)}}[\log(1-D(G(z)))],
\end{eqnarray}
where \textit{z} and \textit{x} denote a random noise vector and a real image sampled from the noise $P_z(z)$ and real data distribution $P_\mathrm{data}(x)$, respectively. Recently, the GAN has been applied to several semantic inpainting techniques~\cite{yu2018generative, pathak2016context, iizuka2017globally} to fill the holes naturally.

\begin{table}[t]
\caption{Detailed architecture of encoding network. }
	\footnotesize
	\begin{center}
		\begin{tabular}{ccccc}
			\hline
			Type & Kernel & Dilation & Stride & Outputs \\
			\hline\hline
			Convolution & $ 5 \times 5 $ & 1 & $ 1 \times 1 $ & 32 \\
			Convolution & $ 3 \times 3 $ & 1 & $ 2 \times 2 $ & 64 \\
			Convolution & $ 3 \times 3 $ & 1 & $ 1 \times 1 $ & 64 \\
			Convolution & $ 3 \times 3 $ & 1 & $ 2 \times 2 $ & 128 \\
			Convolution & $ 3 \times 3 $ & 1 & $ 1 \times 1 $ & 128 \\
			Convolution & $ 3 \times 3 $ & 1 & $ 2 \times 2 $ & 256 \\
			Dilated & \multirow{2}*{$ 3 \times 3 $} & \multirow{2}*{2} & \multirow{2}*{$ 1 \times 1 $} & \multirow{2}*{256} \\
			convolution & & & &\\
			Dilated & \multirow{2}*{$ 3 \times 3 $} & \multirow{2}*{4} & \multirow{2}*{$ 1 \times 1 $} & \multirow{2}*{256} \\
			convolution & & & &\\
			Dilated & \multirow{2}*{$ 3 \times 3 $} & \multirow{2}*{8} & \multirow{2}*{$ 1 \times 1 $} & \multirow{2}*{256} \\
			convolution & & & &\\
			Dilated & \multirow{2}*{$ 3 \times 3 $} & \multirow{2}*{1} & \multirow{2}*{$ 1 \times 1 $} & \multirow{2}*{256} \\
			convolution & & & &\\
			\hline
		\end{tabular}
	\end{center}
	\label{table:table2}
\end{table}


\subsection{Coarse-to-fine network}
\label{sec3.2}
In~\cite{yu2018generative, yu2018free}, a two-stage network called a coarse-to-fine network, which separately conducts a couple of tasks, is proposed. The coarse-to-fine network first generates an initial coarse prediction, $i.e.$ coarse result, using the coarse network, and then refines the results by encoding features from the coarse result with the refinement network. To refine the coarse prediction effectively, they introduced CAM that generates patches of the hole region using features from distant background patches. As depicted in Fig.~\ref{fig:fig1}, the CAM divides the input feature maps into a target foreground and its surrounding background, and extracts $3\times3$ patches. Then, the similarity score $s_{(x,y),(x',y')}$ between the foreground patch at $(x,y)$, $f_{x,y}$, and the background patch at $(x',y')$, $b_{x',y'}$ is computed by using the normalized inner product (cosine similarity), which is expressed as follows:

\begin{eqnarray}
    \label{eq:cossim}
    s_{(x,y),(x',y')} =\left\langle{\frac{f_{x,y}}{\Vert f_{x,y}\Vert}, \frac{b_{x',y'}}{\Vert b_{x',y'}\Vert}}\right\rangle, \\
    s^*_{(x,y),(x',y')} = \mathrm{softmax}(\lambda s_{(x,y),(x',y')}),
\end{eqnarray}
where $\lambda$ is a hyper-parameter for scaled \textit{softmax}. By weighted sum of background patches using $s^*_{(x,y),(x',y')}$ as weights, the CAM rebuilds features of foreground regions, $i.e.$ reconstructed feature. The CAM effectively learns where to borrow or copy the feature information from the background region for the unknown foreground regions, but it requires the coarse result to explicitly attend on related features at distant spatial locations. 


\begin{table}[t]
\caption{Detailed architecture of the decoding network. The output layer consists of a convolution layer clipped value to the [-1, 1].}
	\footnotesize
	\begin{center}
		\begin{tabular}{ccccc}
			\hline
			Type & Kernel & Dilation & Stride & Outputs \\
			\hline\hline
			Convolution $ \times 2 $& $ 3 \times 3 $ & 1 & $ 1 \times 1 $ & 128 \\
			Upsample\scriptsize{ $( \times 2 \uparrow)$} & - & - & -& - \\
			Convolution $ \times 2 $& $ 3 \times 3 $ & 1 & $ 1 \times 1 $ & 64 \\
			Upsample\scriptsize{ $( \times 2 \uparrow)$} & - & - & -& - \\
			Convolution $ \times 2 $ & $ 3 \times 3 $ & 1 & $ 1 \times 1 $ & 32 \\
			Upsample\scriptsize{ $( \times 2 \uparrow)$} & - & - & -& - \\
			Convolution $ \times 2 $ & $ 3 \times 3 $ & 1 & $ 1 \times 1 $ & 16 \\
			Convolution  & \multirow{2}*{$ 3 \times 3 $} & \multirow{2}*{1} & \multirow{2}*{$ 1 \times 1 $} & \multirow{2}*{3} \\
			(Output) &&&&\\
			\hline
		\end{tabular}
	\end{center}
	\label{table:table3}
	\vspace{-0.1cm}
\end{table}


To justify this assumption, we conducted experiments that measure the performance of coarse-to-fine network with/without the coarse path. In our experiments, we trained the refinement network using raw masked images as an input. As shown in Table~\ref{table:table1} and Fig.~\ref{fig: fig2}, the refinement network without the coarse result shows worse results than the full coarse-to-fine network. These results reveal that if the coarse feature of the hole region is not encoded well, the CAM reconstructs features using unrelated feature patches, resulting in inferior results. For instance, as shown in Fig.~\ref{fig: fig2}(d), the refinement network trained without the coarse result produces artifacts such as a wrinkle on the cheek. In other words, the coarse-to-fine network must pass through a two-stage encoder-decoder network which requires massive computational resources. Furthermore, to reduce the operation time of the coarse-to-fine network with another way, we conducted an extra experiment by simplifying the coarse network. In our experiments, we generated the coarse result with low resolution ($64\times64$) and fed it to the refinement network by resizing its resolution to the original size. However, as depicted in Fig.~\ref{fig: fig2}(e) and Table~\ref{table:table1}, the simplified coarse network exhibits worse performance. For instance, as shown in Fig.~\ref{fig: fig2}(e), the simplified coarse network results in asymmetric eyes with defects; the generated right eye has different colors as compared to the left one. These observations indicate that the simplified coarse network can produce the roughly completed image with fast speed but this completed image is not suitable for the refinement network.

\section{Proposed method}
\label{sec4}
\subsection{Architecture of PEPSI}
\label{sec4.1}

As shown in Fig.~\ref{fig:fig3}, PEPSI unifies the stacked networks of the coarse-to-fine network into a single generative network with a single shared encoding network and a parallel decoding network called the coarse and inpainting paths, respectively. The encoding network aims at jointly learning to extract the features from background regions as well as to complete the features of hole regions without the coarse results. As listed in Table~\ref{table:table2}, the encoding network consists of a series of $3\times3$ convolutional layers, except for the first layer which uses a $5\times5$ convolutional layer. To enlarge the receptive field of the encoding network, we utilize multiple dilated convolutional layers with different dilation rates in the last four convolutional layers. 

A parallel decoding network consists of coarse and inpainting paths that share the weight parameters with each other. A detailed architecture of the decoding network is described in Table~\ref{table:table3}. The coarse path produces a roughly completed result from the feature maps obtained via the encoding network, whereas the inpainting path first reconstructs the encoded feature map by using the CAM and produces a higher-quality inpainting result by decoding the reconstructed features. Since two different paths use the same encoded feature maps as their input, this joint learning strategy encourages the encoding network to produce valuable features for two different image generation tasks. To jointly train both paths, we explicitly employ the reconstruction $L_1$ loss to the coarse path, whereas the inpainting path is trained by using both $L_1$ and the adversarial losses. Additional information about the joint learning scheme will be described in Section~\ref{sec4.5}. It should be noted that we employ only the inpainting path during the tests, as depicted in Fig.~\ref{fig:fig0}(c), which substantially reduces the computational complexity. 

In terms of layer implementations in the encoding and decoding networks, PEPSI employs reflection padding for all convolutional layers and uses the exponential linear unit (ELU)~\cite{clevert2015fast} as an activation function, except for the last convolutional layer. In addition, [-1, 1] normalized image with $256\times256$ pixels is employed as an input of PEPSI, and PEPSI produces the output image with the same resolution by clipping the output values into [-1, 1] instead of using the $tanh$ function.




\subsection{Architecture of Diet-PEPSI}
\label{sec4.2}

Although PEPSI effectively reduces the number of convolution operations, it still requires a similar number of network parameters as the coarse-to-fine network. As mentioned in Section~\ref{sec4.1}, PEPSI aggregates the contextual information using a series of dilated convolutional layers, which requires numerous network parameters. The intuitive way to reduce hardware cost is to prune the channels of these layers, but it often yields inferior results in practice. To cope with this problem, we propose novel rate-adaptive dilated convolutional layers that utilize the shared weights but produce dynamic feature maps depending on the given dilation rates. More specifically, to produce rate-specific features, the rate-adaptive dilated convolutional layers alter the shared weights by scaling and shifting differently according to the given dilation rates. Since the rate-adaptive dilated convolutional layers share the weights in every layer, the number of network parameters can be significantly reduced compared with multiple standard dilated convolutional layers. In this subsection, we first introduce how the rate-adaptive dilated convolutional layers produce different feature maps. Then, we explain how the rate-adaptive dilated convolutional layers are applied to PEPSI.

\begin{figure}[t]
\centering
\includegraphics[width=0.9\linewidth]{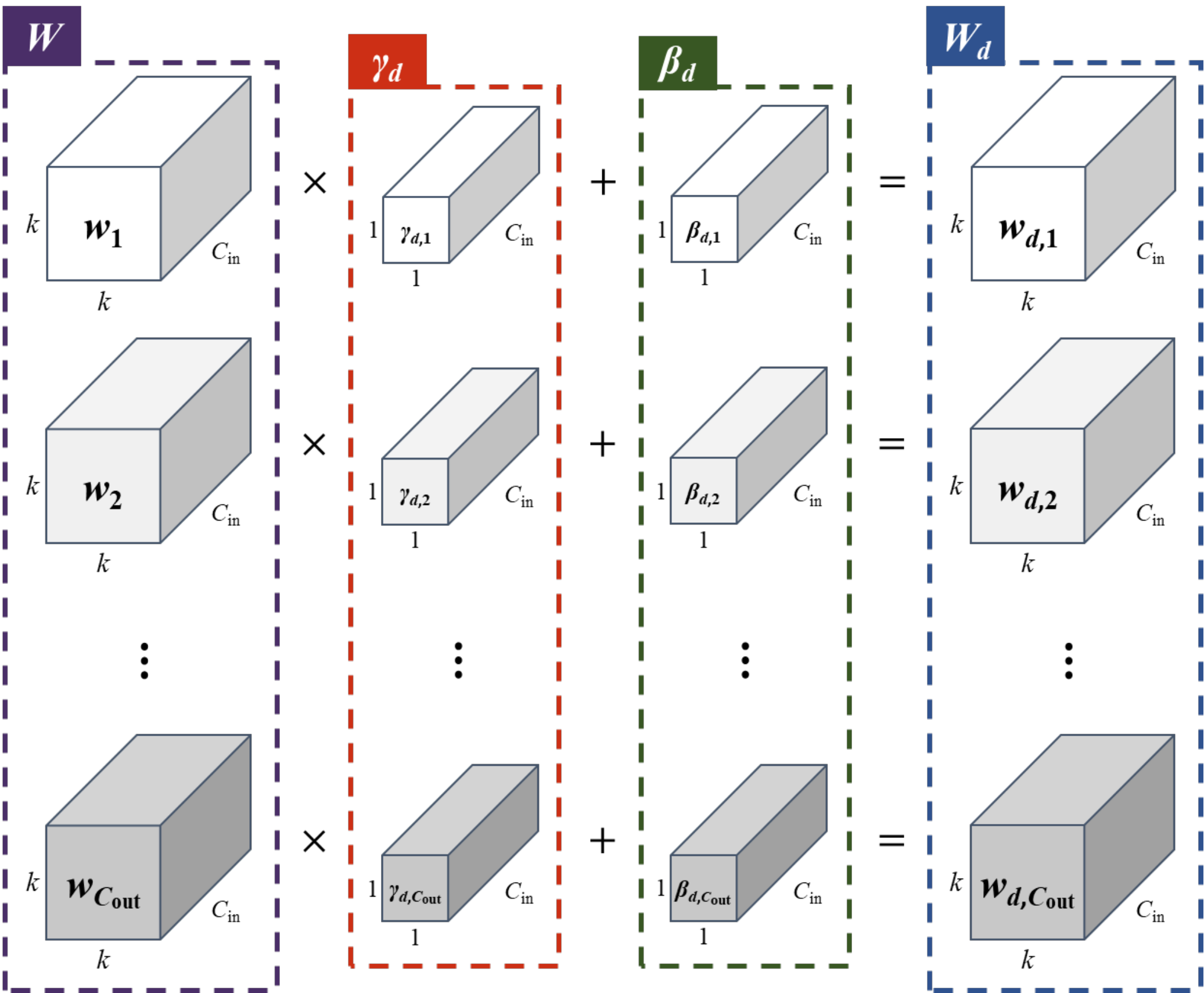}
\caption{Rate-adaptive scaling and shifting operations. $\beta_d$ and $\gamma_d$ have different values depending on the given rate. Tensor broadcasting is included in scaling and shifting operations.} 
\label{fig:fig4_weight}
\vspace{+0.0cm}
\end{figure}

\begin{figure*}[!ht]
\centering
\includegraphics[width=0.9\textwidth]{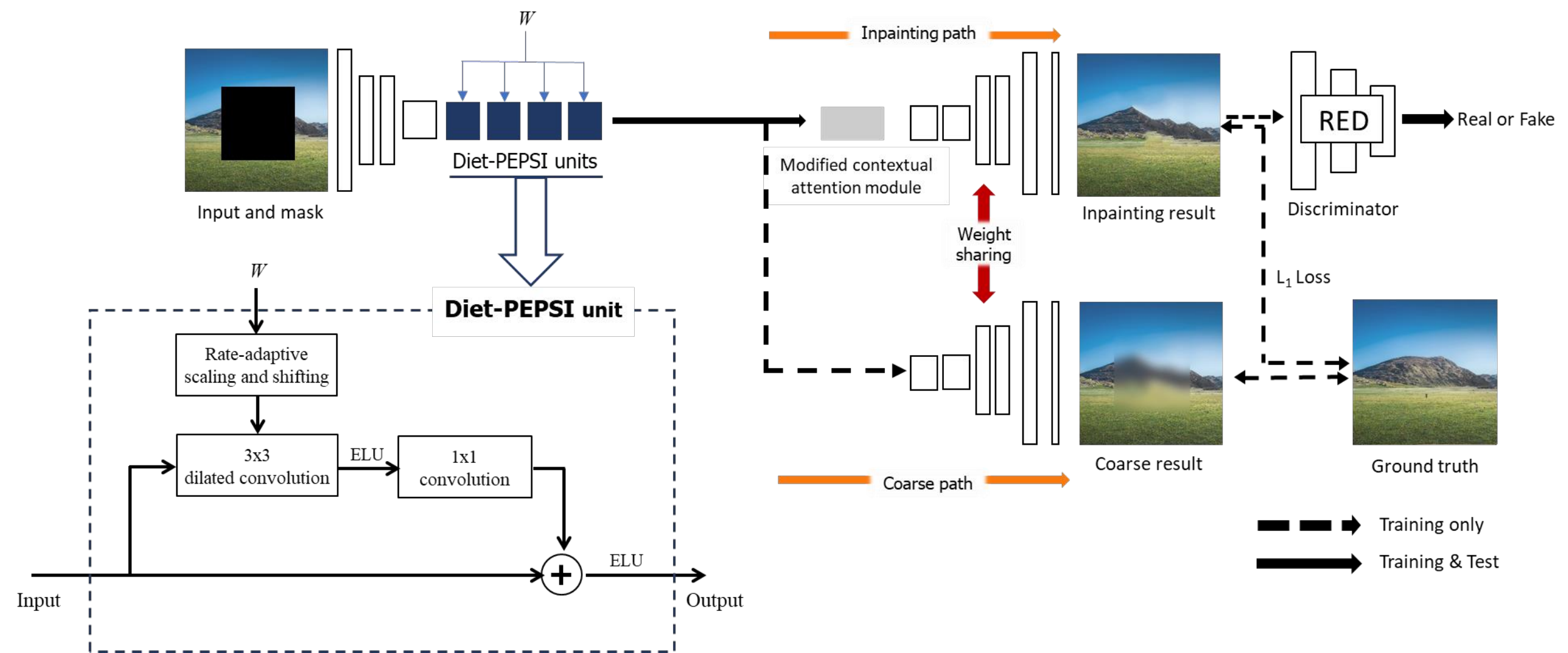}
\caption{Architecture of Diet-PEPSI. We replace the multiple dilated convolutional layers with DPUs. In the DPUs, rate-adaptive convolution layers share their weights whereas the $1~\times1$ standard convolutional layers do not share their weights.}
\label{fig:fig4}
\vspace{-0.3cm}
\end{figure*}

In general, the weights of the convolutional layer are considered as a four-dimensional tensor $W \in \mathbb{R}^{k\times k\times C_{\mathrm{in}} \times C_{\mathrm{out}}}$, where $k$ is the kernel-size, while $C_{\mathrm{in}}$ and $C_{\mathrm{out}}$ are the number of input and output channels, respectively. In other words, as shown in Fig.~\ref{fig:fig4_weight}, the weights in each convolutional layer can be represented as $C_{\mathrm{out}}$ filters with $C_{\mathrm{in}}$ channels, ~\textit{i.e.} $w_i,\{i=1,\dots , C_{\mathrm{out}}\}$. To produce different features according to the given dilation rates, we modulate $W$ using the learned scale $\gamma_d \in \mathbb{R}^{1\times 1\times C_{\mathrm{in}} \times C_{\mathrm{out}}}$ and bias $\beta_d \in \mathbb{R}^{1\times 1\times C_{\mathrm{in}} \times C_{\mathrm{out}}}$ parameters, where \textit{d} indicates the dilation rate; $\gamma_d$ and $\beta_d$ are learned separately depending on the given dilation rate. This modulating process can be expressed as follows:

\begin{eqnarray}
\label{eqn:W_module}
    W_d=\gamma_d \cdot W+\beta_d,
\end{eqnarray}
where $W_d \in \mathbb{R}^{k\times k\times C_{\mathrm{in}} \times C_{\mathrm{out}}}$ represents the rate-adaptively modified weights. Note that tensor broadcasting is included in (4). Using these scaling and shifting processes, the common weights $W$ can be specialized to the desired dilation rate using a small number of parameters. 

\begin{figure}[t]
\centering
\includegraphics[width=0.9\linewidth]{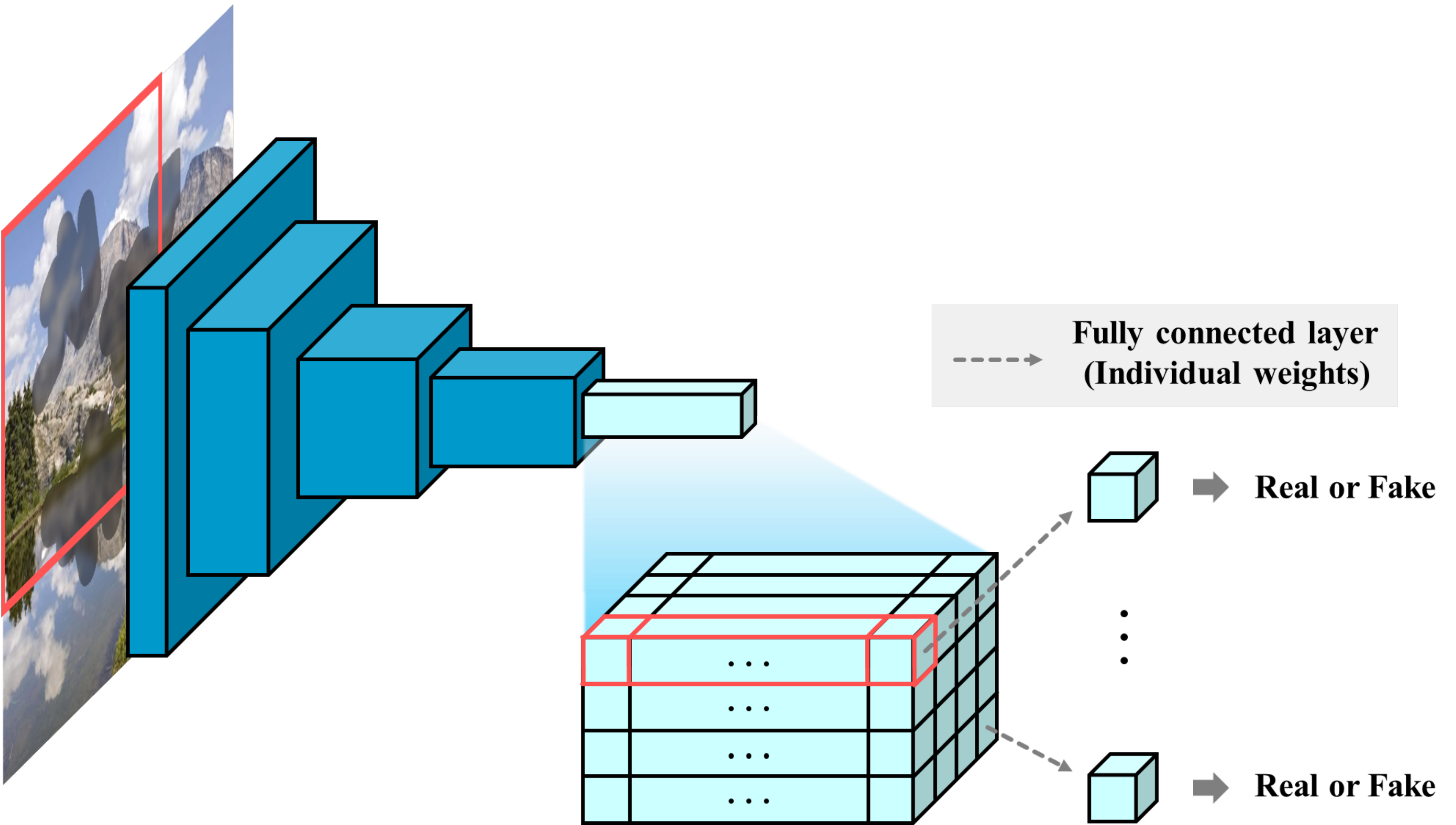}
\caption{Overview of the RED. In the last layer, each pixel employs the fully connected layer with different weights. It aims to classify hole regions that may appear in any region with any sizes in an image.}
\label{fig:fig5}
\vspace{+0.0cm}
\end{figure}

To demonstrate how $W_d$ can generate different feature maps depending on the given dilation rate, we analyze the computational process in the rate-adaptive dilated convolutional layer. The output of this convolutional layer $y$ is formulated as follows:

\begin{eqnarray}
\label{eq:W_demon}
y=x\otimes (\gamma_d W+\beta_d)=x\otimes \gamma_d W+x\otimes \beta_d,
\end{eqnarray}
where $x$ and $\otimes$ indicate the input and convolution operation, respectively. The first term $x\otimes \gamma_d W$ represents a \textit{scaling process} which produces the features that are scaled differently according to the given dilation rate, whereas the second term $x\otimes \beta_d$ is a \textit{projection process} which derives the rate-specific features by projecting $x$ into $\beta_d$. In other words, even though the same features are used as the input of the rate-adaptive convolutional layer, this layer can produce different features depending on the given dilation rates.

\begin{table}[t]
\caption{Detailed architecture of RED. After each convolution layer, except last one, there is a leaky-ReLU as the activation function. Every layer is normalized by a spectral normalization. FC* indicates the fully-connected layer which employs pixel-wise different weights for convolution operation.}
	\footnotesize
	\begin{center}
		\begin{tabular}{cccc}
			\hline
			Type & Kernel & Stride & Outputs \\
			\hline\hline
			Convolution & $ 5 \times 5 $ & $ 2 \times 2 $ & 64 \\
			Convolution & $ 5 \times 5 $ & $ 2 \times 2 $ & 128 \\
			Convolution & $ 5 \times 5 $ & $ 2 \times 2 $ & 256 \\
			Convolution & $ 5 \times 5 $ & $ 2 \times 2 $ & 256 \\
			Convolution & $ 5 \times 5 $ & $ 2 \times 2 $ & 256 \\
			Convolution & $ 5 \times 5 $ & $ 2 \times 2 $ & 512 \\
			FC* & $ 1 \times 1 $ & $ 1 \times 1 $ & 1 \\
			\hline
		\end{tabular}
	\end{center}
	\label{table:table4}
	\vspace{-0.3cm}
\end{table}

Using the rate-adaptive convolutional layers, in this study, we propose a novel lightweight model of PEPSI called Diet-PEPSI, which significantly reduces the network parameters while preserving the inpainting performance. In Diet-PEPSI, as shown in Fig.~\ref{fig:fig4}, we replace the standard dilated convolution layers of PEPSI with residual blocks, \textit{i.e.} DPUs, which consist of a $3~\times3$ rate-adaptive dilated convolutional layer and a $1~\times1$ standard convolutional layer. By increasing the dilation rate, the DPUs can cover the same size of the receptive field with PEPSI. While the standard dilated convolutional layers need $3 \times 3 \times C_{\mathrm{in}} \times C_{\mathrm{out}} \times n$ network parameters, the DPUs require $\left(9+3n\right) \times C_{\mathrm{in}} \times C_{\mathrm{out}}$ network parameters where $n$ indicates the number of DPUs or dilated convolutional layers. Thus, when $n$ is larger than one, DPUs require fewer parameters than the multiple dilated convolutional layers. We will empirically demonstrate the validity of DPUs in Section~\ref{sec5.2}. 

\subsection{Region Ensemble Discriminator (RED)}
\label{sec4.3}

Traditional image inpainting networks~\cite{yu2018generative} utilized both global and local discriminators to determine whether an image has been completed consistently. However, the local discriminator can only handle the hole region with a fixed-size square shape. Thus, it is difficult to employ the local discriminator to train the inpainting network for irregular holes. To solve this problem, we propose a RED inspired by the region ensemble network~\cite{guo2017region} which detects a target object appearing anywhere in the image by individually handling multiple feature regions. As described in Fig.~\ref{fig:fig5} and Table~\ref{table:table4}, six strided convolutions with a kernel size of $5\times5$ and stride 2 are stacked to captures the feature of the whole image. Then, we adopt an individual fully-connected layer on each pixel in the last convolutional layer to individually differentiate that each block is real or fake. In other words, we conduct the $1~\times1$ convolution operation on the last layer using pixel-wise different weights. It is worth noting that the major difference between RED and existing discriminator~\cite{isola2017image}, called PatchGAN-discriminator, is the last convolutional layer. The PatchGAN-discriminator uses the single regressor in the last convolutional layer, whereas the RED employs individual regressors in each pixel. This approach allows the RED to act as global and local discriminator simultaneously. The effectiveness of the RED will be revealed in Section~\ref{sec5.2}.

\subsection{Modified CAM}
\label{sec4.4}
As mentioned in~\ref{sec3.2}, the conventional CAM~\cite{yu2018generative} uses the cosine similarity to measure the similarity scores between foreground and background feature patches. However, in Eq.~\ref{eq:cossim}, since the magnitudes of foreground and background patches, $i.e.$ $f_{x,y}$ and $b_{x',y'}$, are ignored, this approach can result in the distortion of the semantic feature representation. To alleviate this problem, we propose a modified CAM which utilizes the Euclidean distance to measure the similarity scores $(d_{(x,y),(x',y')})$ without the normalization procedure. In the modified CAM, we apply the Euclidean distance instead of the cosine similarity to compute $(d_{(x,y),(x',y')})$. Since the Euclidean distance considers the angle between two vectors of feature patches and their magnitudes simultaneously, it is more appropriate for reconstructing the feature patch. However, since the range of the Euclidean distance is $[0, \infty)$, it is difficult to be directly applied to the \textit{softmax}. To cope with this problem, we define the truncated distance similarity score $\widetilde{d}_{(x,y),(x',y')}$ as follows: 

\begin{eqnarray}
\widetilde{d}_{(x,y),(x',y')} = \tanh{(-(\frac{d_{(x,y),(x',y')}-m(d_{(x,y),(x',y')})}{\sigma (d_{(x,y),(x',y')})}))},\nonumber\\
\label{eq:distsim}
\end{eqnarray}
where $d_{(x,y),(x',y')} = \Vert f_{x,y}-b_{x',y'}\Vert$. Since $\widetilde{d}_{(x,y),(x',y')}$ has limited values within $[-1, 1]$, it operates like a threshold which sorts out the distance scores less than the mean value. In other words, $\widetilde{d}_{(x,y),(x',y')}$ supports to divide the background patches into two groups that may or may not be related to the foreground patch. By using $\widetilde{d}_{(x,y),(x',y')}$, the modified CAM weighs them via scaled softmax and reconstructs the foreground patch using the weighted sum of background ones at last like the conventional CAM. The superiority of the modified CAM will be explained in Section~\ref{sec5.2}. 

\subsection{Loss function}
\label{sec4.5}

To train PEPSI and Diet-PEPSI, we jointly optimize two different paths: the inpainting path and the coarse path. For the inpainting path, we employ the GAN~\cite{goodfellow2014generative} optimization framework in Eq.~\ref{eq:gan}, which is described in Section~\ref{sec3.1}. To avoid the gradient vanishing problem in the generator, inspired by~\cite{zhang2018self}, we employ the \textit{hinge} version of the adversarial loss instead of Eq.~\ref{eq:gan}, which is expressed as follows:

\begin{eqnarray}
L_G = -E_{x\sim P_{X_\mathrm{i}}}[D(x)],
\label{eq:gen2}
\end{eqnarray}
\begin{eqnarray}
\lefteqn{L_{D}  =  E_{x\sim P_Y}[\max(0, 1 - D(x))]}  \nonumber \\
&&\qquad\qquad\qquad{}+ E_{x\sim P_{X_\mathrm{i}}}[\max (0, 1 + D(x))],
\label{eq:disc2}
\end{eqnarray}
where $P_{X_\mathrm{i}}$ and $P_Y$ denote the data distributions of inpainting results and input images, respectively. It is worth noting that we apply the spectral normalization~\cite{miyato2018spectral} to all layers in the RED to further stabilize the training of GANs. Since the goal of the inpainting path is not to produce the hole regions naturally but also to recover the missing part of the original image accurately, we add a strong constraint using $L_1$ norm to Eq.~\ref{eq:gen2} as follows:

\begin{eqnarray}
L_G = \frac{\lambda_\mathrm{i}}{N} \displaystyle\sum_{n=1}^{N}\Vert X_\mathrm{i}^{(n)} - Y^{(n)} \Vert_1 - \lambda_{\mathrm{adv}}E_{x\sim P_{X_\mathrm{i}}}[D(x)],
\end{eqnarray}
where $X_\mathrm{i}^{(n)}$ and $Y^{(n)}$ represent the $n$-th image pair of the generated image through the inpainting path and its corresponding original image in a mini-batch, respectively, $N$ is the number of image pairs in a mini-batch, and $\lambda_\mathrm{i}$ and $\lambda_\mathrm{adv}$ are hyper-parameters which control the relative importance of each loss term. 

On the other hand, the coarse path is designed to accurately restore the missing features for the CAM. Therefore, we simply optimize the coarse path using an $L_1$ loss function which is defined as follows:

\begin{figure}[t]
\centering
\vspace{-0.2cm}
\includegraphics[width=0.85\linewidth]{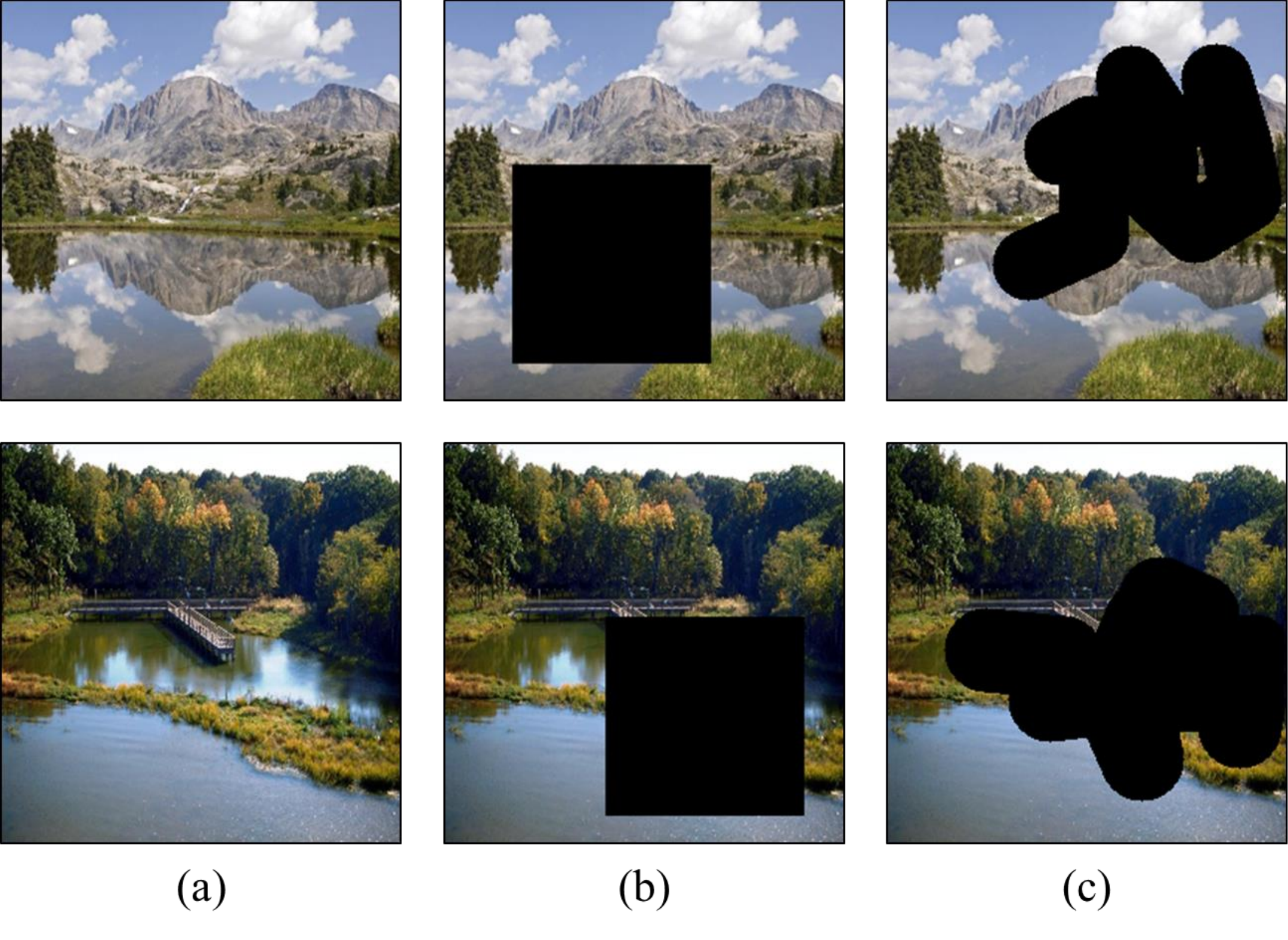}
\vspace{-0.3cm}
\caption{Examples of masked image. (a) Original images. (b) Images with square mask. (c) Images with free-form mask.}
\label{fig:fig6}
\vspace{-0.3cm}
\end{figure}

\begin{figure*}[!ht]
\centering
\includegraphics[width=0.9\linewidth]{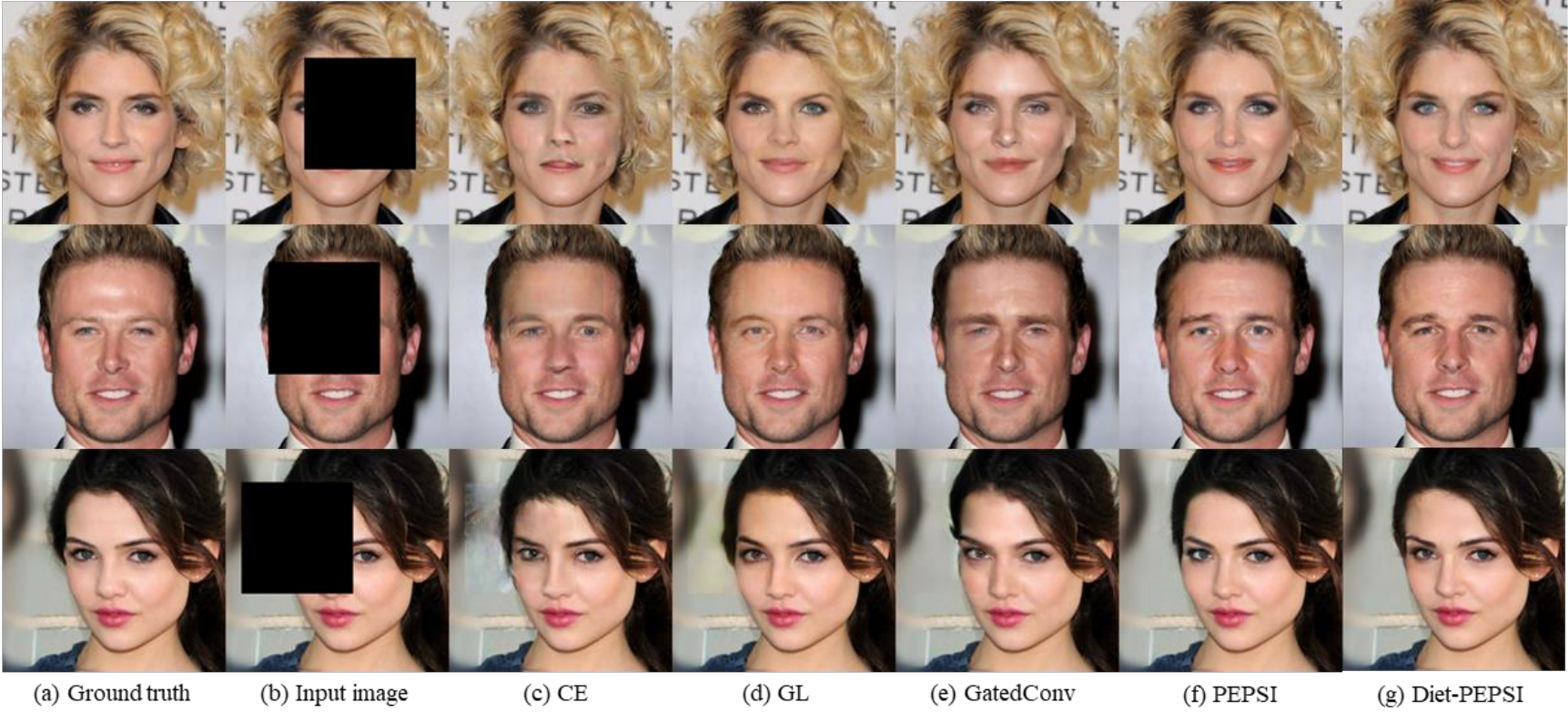}
\caption{Comparison of the proposed and conventional methods on randomly square masked CelebA-HQ datasets. (a) Ground truth. (b) Input image of the network. (c) Results of Context Encoder~\cite{pathak2016context}. (d) Results of Globally-Locally~\cite{iizuka2017globally}. (e) Results of gated convolution~\cite{yu2018free}. (f) Results of PEPSI. (g) Results of Diet-PEPSI.}
\label{fig:fig7}
\vspace{-0cm}
\end{figure*}

\begin{eqnarray}
L_{C} = \frac{1}{N}\displaystyle\sum_{n=1}^{N}\Vert X_\mathrm{c}^{(n)} - Y^{(n)} \Vert_1,
\end{eqnarray}
where $X_\mathrm{c}^{(n)}$ are the $n$-th image pair of the generated image via the coarse path in a mini-batch. Finally, we define the total loss function of the generative network of PEPSI and Diet-PEPSI as follows:

\begin{eqnarray}
L_{\mathrm{total}} = L_{G} + \lambda_\mathrm{c} (1- \frac{k}{k_\mathrm{max}}) L_{C},
\end{eqnarray}
where $\lambda_{c}$ is a hyper-parameter controlling the contributions from each loss term, and $k$ and $k_\mathrm{max}$ represent the iteration of the learning procedure and the maximum number of iterations, respectively. As the training progresses, we gradually decrease the contribution of the $L_C$ for the decoding network to focus on the inpainting path. More specifically, as the training progresses, $(1-k/k_\mathrm{max})$ becomes zero, which results in reducing the contribution of $L_C$. 

\begin{figure*}[!ht]
\centering
\includegraphics[width=0.9\linewidth]{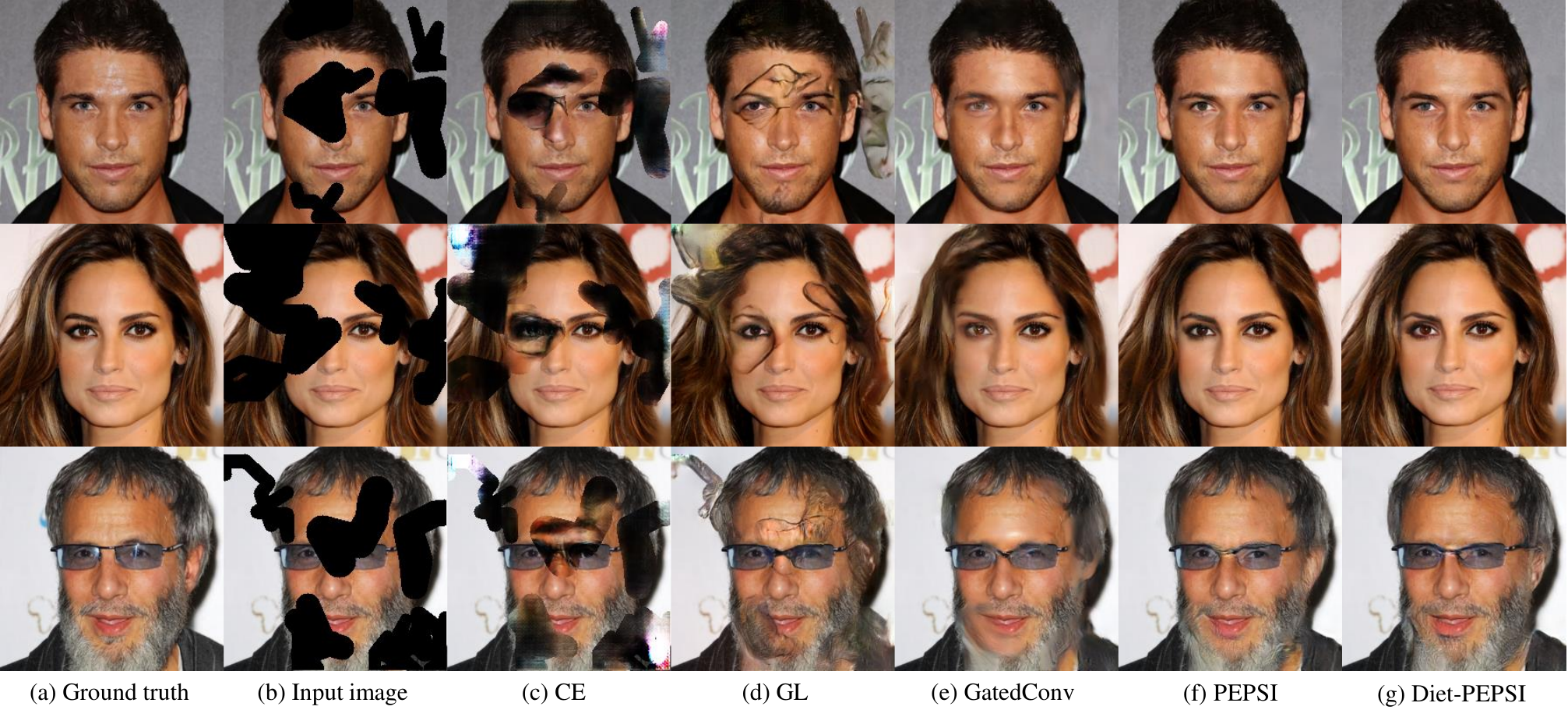}
\caption{Comparison of the proposed and conventional methods on free-form masked CelebA-HQ datasets. (a) Ground truth. (b) Input image of the network. (c) Results of Context Encoder~\cite{pathak2016context}. (d) Results of Globally-Locally~\cite{iizuka2017globally}. (e) Results of gated convolution~\cite{yu2018free}. (f) Results of PEPSI. (g) Results of Diet-PEPSI.}
\label{fig:fig8}
\vspace{-0.2cm}
\end{figure*}

\section{Experiments}
\label{sec5}
\subsection{Implementation details}
\label{sec5.1}

\textbf{Free-Form Mask} As shown in Fig.~\ref{fig:fig6}(b), existing image inpainting methods~\cite{yu2018generative, iizuka2017globally, pathak2016context} usually adopt the regular mask, \textit{e.g.} hole region with rectangular shape, which indicates the background regions during the training procedure. However, the networks trained with the regular mask often exhibit weak performance on inpainting the hole with irregular shape and result in visual artifacts such as color discrepancy and blurriness. To address this problem, as depicted in Fig.~\ref{fig:fig6}(c), Yu~\textit{et al.}~\cite{yu2018free} adopted the free-form mask algorithm during the training procedure, which automatically generates multiple random free-form holes with variable numbers, sizes, shapes, and locations randomly sampled at every iteration. More specifically, this algorithm first produces the free-form mask by drawing multiple different lines and erasing pixels closer than the arbitrary distance from these lines. For a fair comparison, in our experiments, we employed the same free-form mask generation algorithm for training PEPSI and Diet-PEPSI.\\ 

\begin{figure*}[!ht]
\centering
\includegraphics[width=0.95\linewidth]{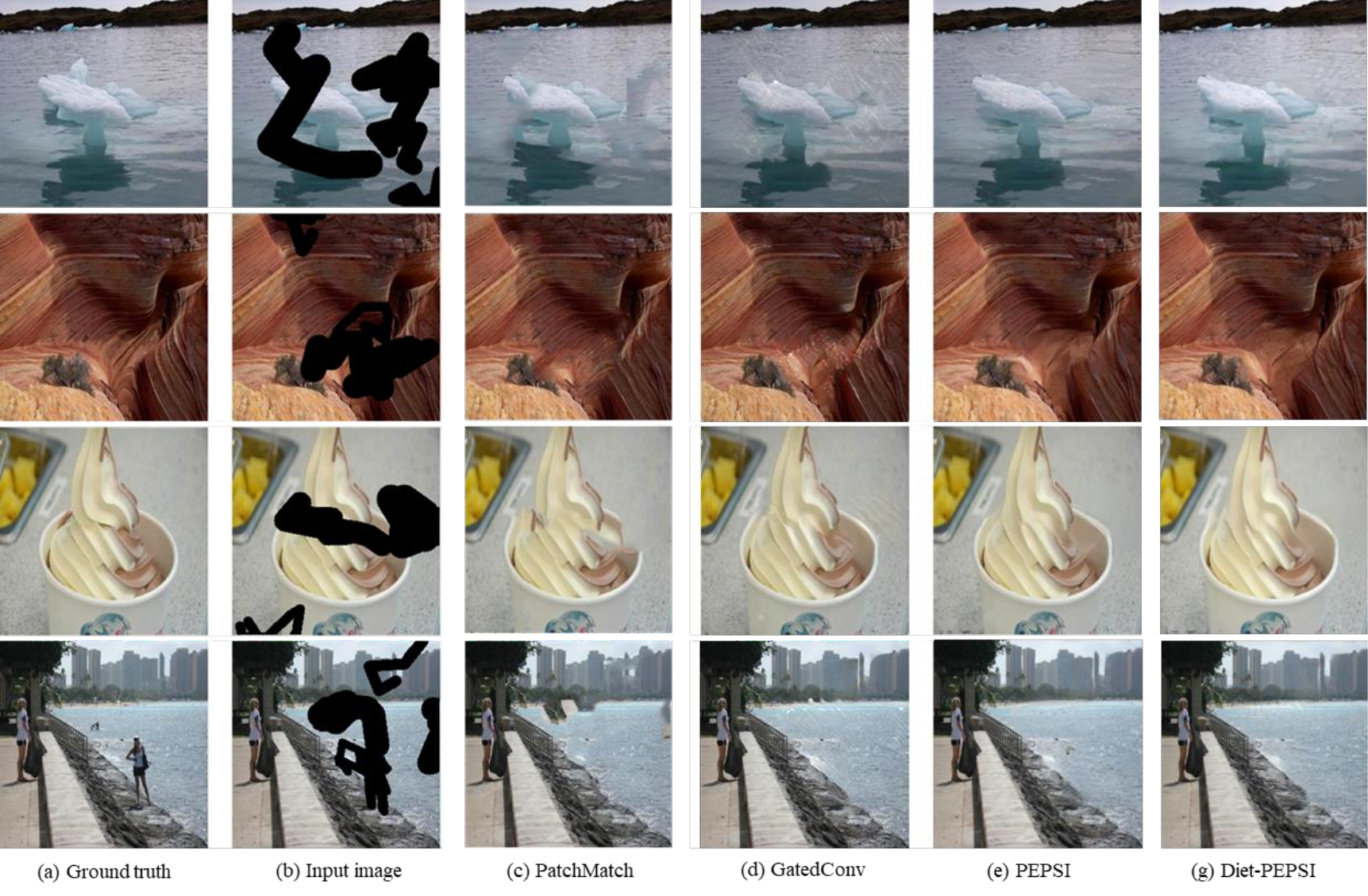}
\caption{Comparison of the proposed and conventional methods on Place2 dataset. (a) Ground truth. (b) Input image of the network. (c) Results of the non-generative method, PatchMatch~\cite{barnes2009patchmatch}. (d) Results of GatedConv~\cite{yu2018free}. (e) Results of PEPSI. (f) Results of Diet-PEPSI.}
\label{fig:fig9}
\vspace{-0.2cm}
\end{figure*}


\begin{table*}[t]
\caption{Results of global and local PSNRs, SSIM, and operation time with square and free-formed masks on CelebA-HQ dataset.}
\vspace{+0.1cm}
\begin{center}
\begin{tabular}{c c c c | c c c c c}
\hline
& \multicolumn{3}{c}{Square mask} & \multicolumn{3}{c}{Free-form mask} & \multirow{3}*{Time~(ms)} & Number of\\
\cline{2-7}
 \multirow{2}*{Method} & \multicolumn{2}{c}{PSNR} & \multirow{2}*{SSIM} & \multicolumn{2}{c}{PSNR} & \multirow{2}*{SSIM} & & Network \\ \cline{2-3} \cline{5-6} 
  & Local & Global & & Local & Global & & & Parameters \\
\hline\hline
CE\cite{pathak2016context}&17.7 &23.7 &0.872 &9.7 &16.3 &0.794 & \bf{5.8} & 5.1M\\
GL\cite{iizuka2017globally}&\underline{19.4} &25.0 &0.896&15.1 &21.5 &0.843 & 39.4 & 5.8M \\
GCA\cite{yu2018generative}&19.0 &24.9 &\underline{0.898}&12.4 &18.9 &0.798 & 22.5 & 2.9M\\
GatedConv\cite{yu2018free}&18.7 &24.7 &0.895&21.2 &27.8 &0.925 & 21.4 & 4.1M\\
PEPSI      &\bf{19.5} &\bf{25.6} &\bf{0.901}&\bf{22.0} &\bf{28.6} &\bf{0.929} & \multirow{2}*{\underline{9.2}} & \multirow{2}*{3.5M}\\
PEPSI w/o coarse path &19.2&25.2&0.894&21.6&28.2&0.923& \\
Diet-PEPSI &\underline{19.4} &\underline{25.5} &\underline{0.898}&\bf{22.0} &\underline{28.5} &\underline{0.928} & 10.9 & 2.5M\\

\hline
\end{tabular}
\end{center}
\vspace{+0.1cm}
\label{table:table5}
\end{table*}

\textbf{Training Procedure} PEPSI and Diet-PEPSI were trained for one million iterations using a batch size of eight in an end-to-end manner. Because the parameters in PEPSI and Diet-PEPSI can be differentiated, we performed an optimization using the Adam optimizer~\cite{kingma2014adam} and set the parameters of Adam optimizers $\beta _1$ and $\beta _2$ to 0.5 and 0.9, respectively. Motivated by~\cite{heusel2017gans}, we applied the two-timescale update rule (TTUR) where the learning rates of the discriminator and generator were ${4}\times{10}^{-4}$ and ${1}\times{10}^{-4}$, respectively. In addition, we reduced the learning rate to 1/10 after 0.9 million iterations. The hyper-parameters of the proposed method were set to $\lambda_\mathrm{i} = 10$, $\lambda_\mathrm{c} = 5$, and $\lambda_{\mathrm{adv}} = 0.1$. Our experiments were conducted using CPU Intel(R) Xeon(R) CPU E3-1245 v5 and GPU TITAN X (Pascal), and implemented in TensorFlow v1.8.

For our experiments, we used the CelebA-HQ~\cite{liu2015deep, karras2017progressive}, ImageNet~\cite{krizhevsky2012imagenet}, and Place2~\cite{zhou2018places} datasets. More specifically, in the CelebA-HQ dataset, we randomly sampled the 27,000 images as a training set and 3,000 ones as a test set. We also trained the network with all images in the ImageNet dataset and tested it on Place2 dataset to measure the performance of trained deep learning models on other datasets; these experiments were conducted to confirm the generalization ability of the proposed method. To demonstrate the superiority of PEPSI and Diet-PEPSI, in addition, we compared their qualitative, quantitative, operation speeds, and number of network parameters with those of the conventional generative methods: context encoders (CE)~\cite{pathak2016context}, globally and locally completion network (GL)~\cite{iizuka2017globally}, generator with contextual attention (GCA)~\cite{yu2018generative}, and generator with gated convolution (GatedConv)~\cite{yu2018free}. 


\subsection{Performance Evaluation}
\label{sec5.2}

\textbf{Qualitative Comparison}
To reveal the superiority of PEPSI and Diet-PEPSI, we compared the qualitative performance of the proposed methods with those of the conventional generative methods using images with the squared mask and free-form mask. In our experiments, we implemented the conventional methods by following the training procedure in each study. The resultant images with squared mask and free-form mask are described in Figs.~\ref{fig:fig7} and~\ref{fig:fig8}. As shown in Figs.~\ref{fig:fig7} and~\ref{fig:fig8}, CE~\cite{pathak2016context} and GL~\cite{iizuka2017globally} show obvious visual artifacts including blurred or distorted images in the masked region. In particular, these methods show inferior performance when inpainting the free-form mask, which indicates that CE~\cite{pathak2016context} and GL~\cite{iizuka2017globally} cannot be applied to real applications. GatedConv~\cite{yu2018free} exhibits fine performance compared to CE~\cite{pathak2016context} and GL~\cite{iizuka2017globally}, it still suffers from a lack of relevance between the hole and background regions such as symmetry of eyes. Compared to the existing methods, PEPSI shows visually appealing results and high relevance between hole and background regions. In addition, as shown in Fig.~\ref{fig:fig7}(g) and Fig.~\ref{fig:fig8}(g), the output image produced via Diet-PEPSI were comparable to PEPSI while saving a significant number of network parameters. From these results, we confirmed that the proposed methods outperform compared with the conventional methods, while significantly reducing the hardware costs. 

Furthermore, we trained and tested PEPSI and Diet-PEPSI using the challenging datasets, $i.e.$ ImageNet and Place2 datasets, to demonstrate that the proposed methods can be applied to real applications. In this paper, we compared performance of the proposed methods with that of the GatedConv and the non-generative method, called PatchMatch~\cite{barnes2009patchmatch}, which is widely applied to image editing applications. We set the image resolution as $256\times256$. Resultant images are depicted in Fig.~\ref{fig:fig9}. PatchMatch shows visually poor performance especially on the edge of images since it cannot consider the global contexts of the image for inpainting the hole region. GatedConv generates more realistic results without color discrepancy or edge distortion compared to PatchMatch technique. However, it often produces the images with wrong textures as shown in the first and third rows in Fig.~\ref{fig:fig9}. In contrast to the conventional methods, PEPSI and Diet-PEPSI generate the most natural images without artifacts or distortion on various contents and complex scenes. Thus, we confirmed that the proposed method can be applied to the real application for image inpainting.\\

\begin{table}[t]
\caption{Experimental results that further reduce the network parameters using the group convolution technique.}
\footnotesize
\vspace{-0.1cm}
\begin{center}
\begin{tabular}{cccccc}
\hline
 & \multicolumn{2}{c}{Square mask} & \multicolumn{2}{c}{Free-form mask} & Number of \\ \cline{2-5}
  & PSNR & SSIM & PSNR & SSIM & parameters \\
\hline\hline
PEPSI &25.6 &0.901 &28.6 &0.929 & 3.5M\\
Diet-PEPSI &25.5 & 0.898 & 28.5 & 0.928 & 2.5M\\
Diet-PEPSI ($g=2$) & 25.4 & 0.896 & 28.5 & 0.928 & 1.8M\\
Diet-PEPSI ($g=4$) & 25.2 & 0.894 & 28.4 & 0.926 & 1.5M\\
\hline
\end{tabular}
\end{center}
\vspace{-0cm}
\label{table:Diet_additional}
\vspace{-0.3cm}
\end{table}

\textbf{Quantitative Comparison}
In this study, we adopted the two different metrics for quantitative assessment: peak signal-to-noise ratio (PSNR) of the local and global regions, $i.e.$ PSNR of the hole region and the whole image, and structural similarity (SSIM)~\cite{wang2004image} of the whole image. Table~\ref{table:table5} provides the comprehensive performance benchmarks between the proposed methods and conventional ones~\cite{yu2018generative, iizuka2017globally, pathak2016context, yu2018free} in the CelebA-HQ datasets~\cite{karras2017progressive}. As shown in Table~\ref{table:table5}, compared with the proposed methods, CE~\cite{pathak2016context} and GCA~\cite{yu2018generative} shows worse performance on both square mask and free-form mask. GL~\cite{iizuka2017globally} exhibits comparable performance with the proposed methods only in the square mask since it uses an image blending technique as a post-processing. However, it needs additional computation time owing to the post-processing and still suffers from blurred images as shown in Fig.~\ref{fig:fig7}. Also, like the CE and GCA, GL shows poor performance on free-from mask. Because these methods, $i.e.$ CE, GCA, and GL, designed for inpainting the rectangular mask, they could not cover the free-from mask; they could not generalize well on the free-form mask.  

GatedConv~\cite{yu2018free} shows better performance in both square and free-form holes than other existing methods, but it needs some computation time owing to the two stacked generative networks. Compared with conventional methods, PEPSI and Diet-PEPSI show fine performance in both square and free-from masks. In particular, compared with GatedConv, PEPSI and Diet-PEPSI not only exhibit better PSNR and SSIM performances but also require less computational time and hardware costs. In addition, Diet-PEPSI achieves comparable performance with PEPSI while reducing the network parameters almost by 30 percent. Consequently, these observations indicate that the proposed methods can successfully generate inpainting results with high-quality and less hardware costs compared with the conventional inpainting techniques. 

\begin{table}[t]
\caption{Results of global and local PSNRs and SSIM on Place2 dataset.}
\footnotesize
\begin{center}
\begin{tabular}[h]{ccccc}
\hline
 \multirow{2}*{Mask} &\multirow{2}*{Method} & \multicolumn{2}{c}{PSNR} & \multirow{2}*{SSIM} \\ \cline{3-4}
 & & Local & Global & \\
\hline\hline
\multirow{3}*{Square} &GatedConv\cite{yu2018free}&14.2 &20.3 &0.818 \\
&PEPSI     &15.2 &21.2 &0.832 \\
&Diet-PEPSI    & \textbf{15.5} & \textbf{21.5} & \textbf{0.840} \\
\hline
\multirow{3}*{Free-form} &GatedConv\cite{yu2018free}&17.4 &24.0 &0.875 \\
&PEPSI     &18.2 &24.8 &0.882 \\
&Diet-PEPSI     & \textbf{18.7} & \textbf{25.2} & \textbf{0.889} \\
\hline
\end{tabular}
\end{center}
\vspace{-0.1cm}
\label{table:table6}

\end{table}

Moreover, to reveal the effectiveness of the coarse path, we conducted an extra experiment in which PEPSI was trained without using the coarse path learning. The experimental results are described in Table~\ref{table:table5}. PEPSI exhibits the better performance than PEPSI trained without the coarse path in terms of all quantitative metrics. These results demonstrate that the coarse path drives the encoding network to produce missing features properly for the CAM. In other words, the single-stage network structure of PEPSI can overcome the limitation of the two-stage coarse-to-fine network through a joint learning scheme. 


On the other hand, Diet-PEPSI retains the ability of PEPSI while significantly reducing the network parameters as shown in Table~\ref{table:table5}. These results reveal that the DPU with rate-adaptive convolutional layer can replace the standard dilated convolutional layer with a small number of network parameters. To further reduce the hardware costs of Diet-PEPSI, we conducted additional experiments that apply the group convolution technique~\cite{zhang2018shufflenet} to the DPU. In our experiments, we trained Diet-PEPSI by employing the group convolution technique to both layers in the DPU. Note that we utilized the channel shuffling technique between the two convolutional layers of the DPU. As shown in Table~\ref{table:Diet_additional}, even though Diet-PEPSI utilizes a significantly less number of network parameters, it achieves competitive performance with PEPSI as well as shows superior performance compared to other conventional methods. These results confirm that Diet-PEPSI can generate high-quality images with low hardware costs. 
\begin{figure}[t]
\centering
\includegraphics[width=0.9\linewidth]{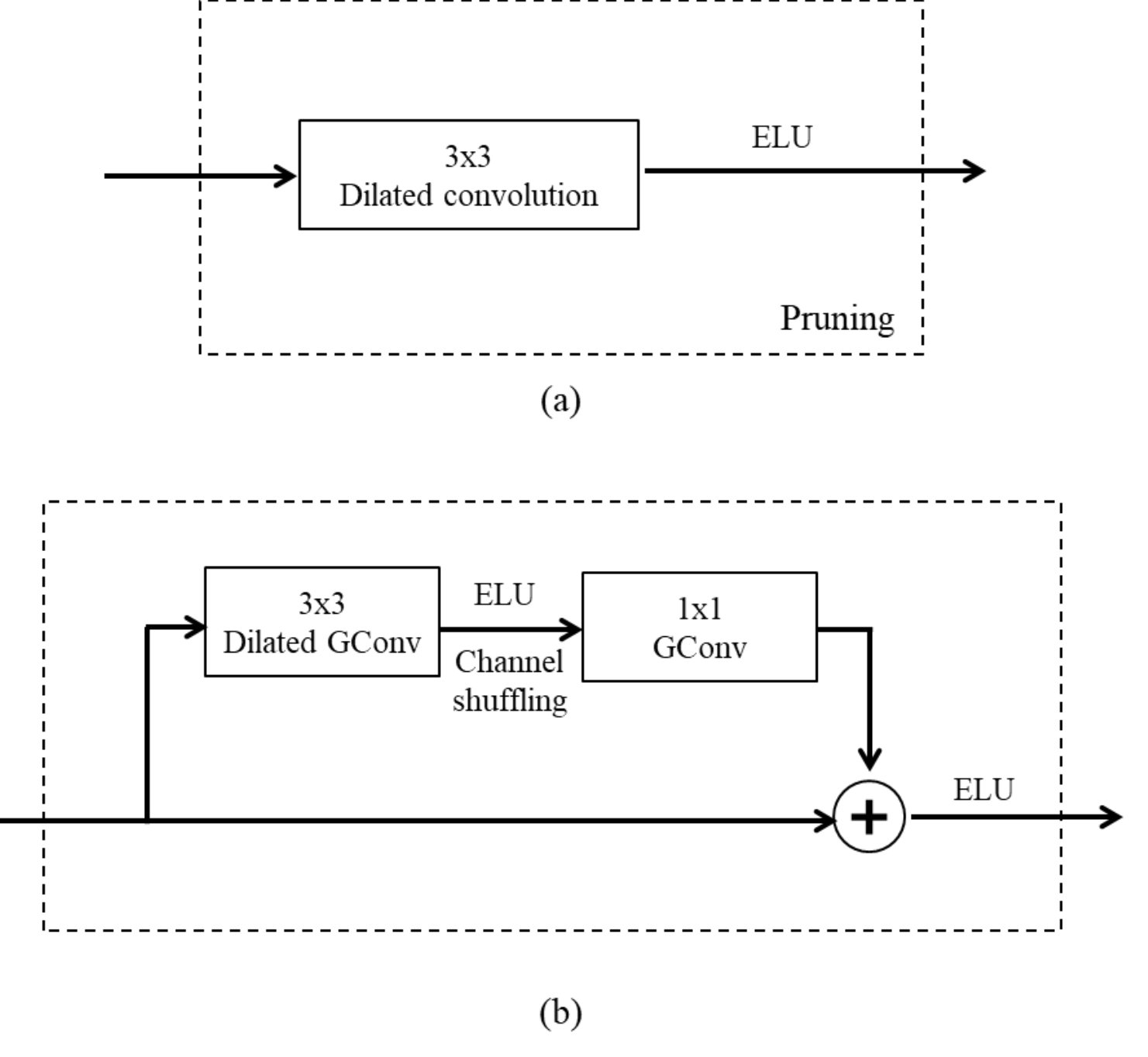}
\vspace{+0.1cm}
\caption{Illustration of techniques to aggregate the global contextual information while reducing the number of parameters. (a) Dilated convolutional layer with pruning channel. (b) Residual block consisting of group convolutional layers.}
\label{fig:fig10}

\end{figure}

\begin{table}[t]
\caption{Experimental results using different lightweight units.}
\footnotesize
\vspace{-0.1cm}
\begin{center}
\begin{tabular}{ccccc}
\hline
 & \multicolumn{2}{c}{Square mask} & \multicolumn{2}{c}{Free-form mask} \\ \cline{2-5}
  & PSNR & SSIM & PSNR & SSIM \\
\hline\hline
Pruning &25.21 &\textbf{0.8961} &28.28 &0.9270 \\
DGC &25.28 & 0.8959 & 28.43 & 0.9270 \\
DPU & \textbf{25.38} & 0.8960 & \textbf{28.53} & \textbf{0.9278}\\
\hline
\end{tabular}
\end{center}
\vspace{-0cm}
\label{table:table7}

\end{table}

To demonstrate the generalization ability of PEPSI and Diet-PEPSI, we conduct another experiment using the challenging datasets, ImageNet~\cite{krizhevsky2012imagenet} and Place2~\cite{zhou2018places}. As mentioned in~\ref{sec5.1}, in our experiments, we trained the network using the ImageNet dataset and tested the trained network on the Place2 dataset. Among the various conventional methods, we selected the GatedConv~\cite{yu2018free}, which exhibits superior performance compared to other conventional methods in CelebA-HQ dataset, as our comparison. As shown in Table~\ref{table:table6}, PEPSI achieves better performance than GatedConv in the Place2 dataset. Furthermore, Diet-PEPSI exhibits superior performance compared to GatedConv and PEPSI. These results indicate that the proposed methods can consistently generate high-quality results using various contents and complex images. \\


\begin{figure}[t]
\centering
\vspace{+0.3cm}
\includegraphics[width=0.95\linewidth]{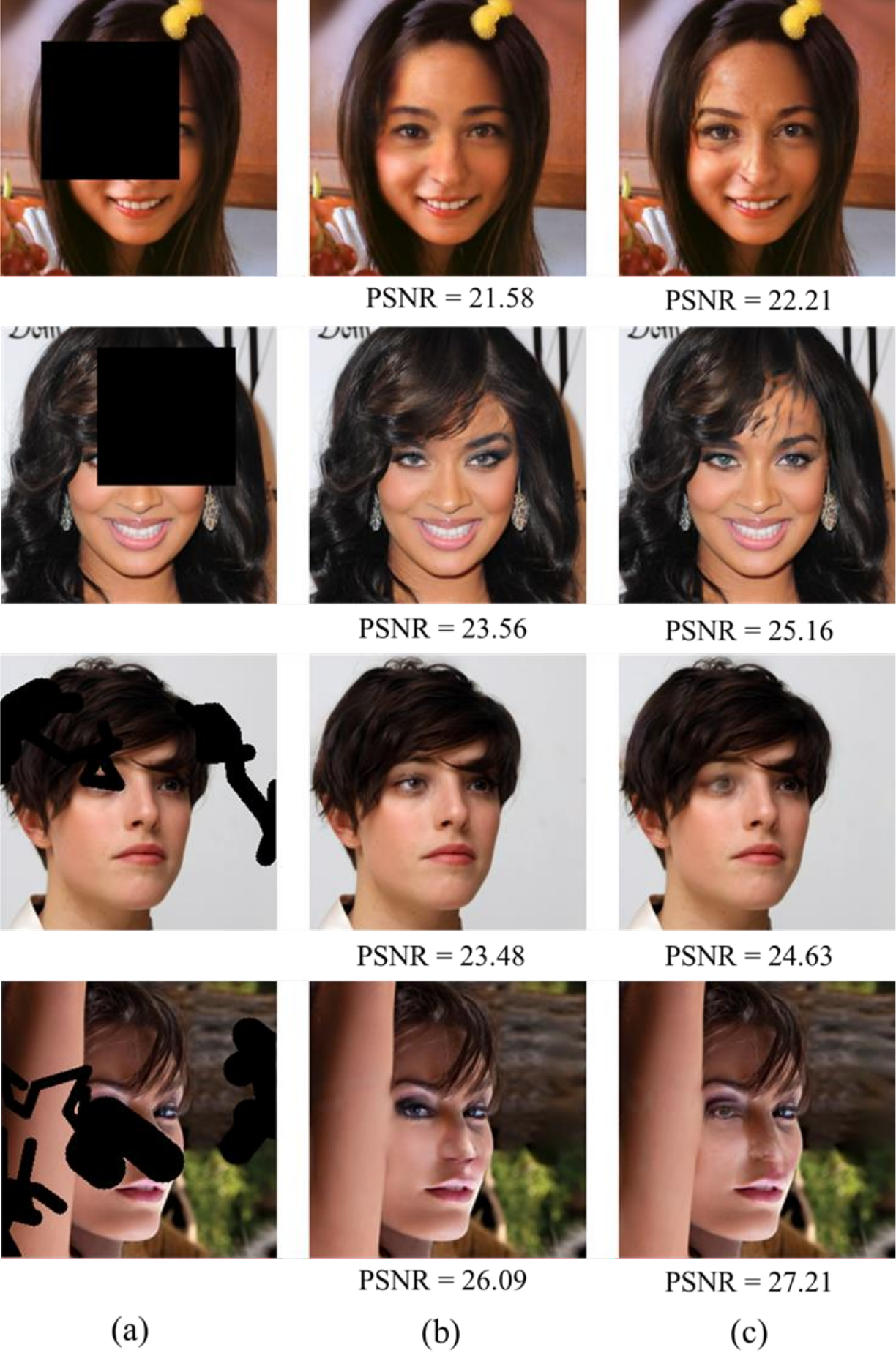}
\vspace{-0.3cm}
\caption{Comparison of RED and SNM-Disc~\cite{yu2018free} on CelebA-HQ datasets. (a) Input image. (b) Results of PEPSI trained with RED. (c) Results of PEPSI trained with SNM-Disc.}
\label{fig:fig11}
\end{figure}


\begin{table}[t]
\caption{Experimental results using different discriminators.}
\footnotesize
\begin{center}
\vspace{+0.3cm}
\begin{tabular}{ccccc}
\hline
 & \multicolumn{2}{c}{Square mask} & \multicolumn{2}{c}{Free-form mask} \\ \cline{2-5}
  & PSNR & SSIM & PSNR & SSIM \\
\hline\hline
SNM-Disc~\cite{yu2018free} & 25.68 & 0.901 & 28.71 & 0.932 \\
RED & 25.57 & 0.901 & 28.59 & 0.929 \\

\hline
\end{tabular}
\end{center}
\vspace{-0cm}
\label{table:table8}
\end{table}


\textbf{DPU analysis}
To demonstrate the ability of the DPU, we conducted additional experiments that reduce the network parameters using different techniques. Fig.~\ref{fig:fig10} shows the models used in our experiments. Figs.~\ref{fig:fig10}(a) and (b) illustrate the convolution layer with a pruning channel and the residual block with dilated group convolutional layer, respectively, which are an intuitive approach to decrease the number of parameters. Note that we employed the residual block with the same architecture as the DPU for a fair comparison. Additionally, we adjusted the pruning channel and the number of groups to make models using an almost similar number of parameters. In our experiments, we set the channels of pruned convolution layers to 113 and the group numbers of the residual block with dilated group convolutional layer to four. The number of groups in the DPU is set to two. As shown in Table~\ref{table:table7}, the pruning strategy shows inferior quantitative scores in terms of PSNR in both square and free-form masks. Although the residual block with group dilated convolutional layer shows slightly better performance compared to the pruning strategy, it is still weak. Compared with these models, the DPU shows superior performance in both square and free-form masks. Therefore, these results confirm that the DPU is suitable to effectively aggregate the global contextual information with a small number of parameters. \\


\textbf{RED analysis}
We demonstrated the superiority of RED by comparing its performance to that of the SNM-discriminator~\cite{yu2018free} (SNM-Disc), which is an extended version of the PatchGAN-discriminator for image inpainting with free-form mask. For fair comparison, we employed each discriminator on the same generator with PEPSI. As shown in Table~\ref{table:table8}, the SNM-Disc exhibits slightly better performance in terms of PSNR and SSIM compared to the RED. However, Fig.~\ref{fig:fig11} shows that the SNM-Disc could not generate a visually plausible image despite having a high PSNR value; PEPSI trained with the SNM-Disc produced the results with visual artifacts such as blurred or distorted images in the masked region. These results indicate that the SNM-Disc cannot effectively compete with the generative networks, which makes the generator mainly focus on minimizing the $L_1$ loss in the objective function of PEPSI. Therefore, even though PEPSI trained with the SNM-Disc exhibits good quantitative performance, it is difficult to apply to the image inpainting in practice.

\begin{figure}[t]
\centering
\vspace{+0.3cm}
\includegraphics[width=0.9\linewidth]{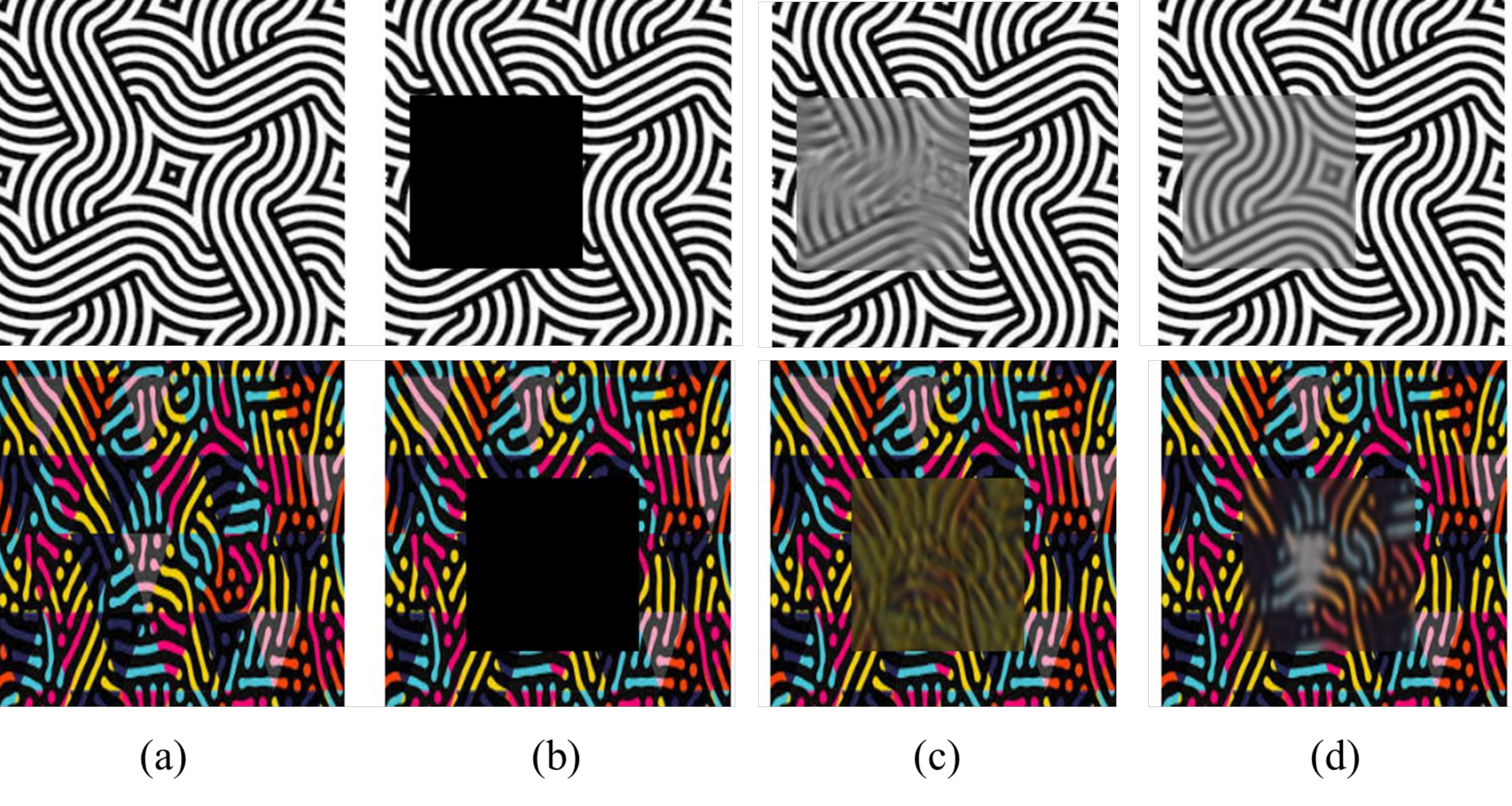}
\vspace{-0.3cm}
\caption{Comparisons of image reconstruction between the cosine similarity and truncated distance similarity. (a) Original images. (b) Masked images. (c) Images reconstructed by using the cosine similarity. (d) Images reconstructed by using the truncated distance similarity.}
\label{fig:fig12}
\end{figure}

\begin{table}[t]
\caption{Performance comparison between cosine similarity and Euclidean distance applying on PEPSI.}
\footnotesize
\begin{center}
\begin{tabular}{ccccc}
\hline
 & \multicolumn{2}{c}{Square mask} & \multicolumn{2}{c}{Free-form mask} \\ \cline{2-3}\cline{4-5}
  & PSNR & SSIM &PSNR &SSIM \\
\hline\hline
Cosine similarity&25.16 &0.8950 &27.95&0.9218 \\
Euclidean distance&25.57 &0.9007 &28.59&0.9293 \\
\hline
\end{tabular}
\end{center}
\label{table:table9}
\end{table}

On the other hand, we investigated the reason why RED could effectively drive the generator to produce visually pleasing inpainting results. The RED follows the inspiration of the region ensemble network~\cite{guo2017region} which classifies objects in any region of the image. Thus, in adversarial learning, the generator attempts to produce every region of the image to be indistinguishable from real images. This procedure further improves the performance of the generator in free-form masks including irregular holes. Thus, we expect that RED can be applied to various image inpainting networks for generating visually plausible images.\\ 

\textbf{Modified CAM analysis}
To demonstrate the validity of modified CAM, we performed toy examples comparing the cosine similarity and truncated distance similarity. We reconstructed the hole region using the weighted sum of existing image patches where the weights, $i.e.$ similarity scores, are computed by using the cosine similarity or truncated Eculidean distance. Fig.~\ref{fig:fig12} shows comparisons of reconstructed images. As depicted in Figs.~\ref{fig:fig12}(c) and (d), images reconstructed by applying the truncated distance similarity can collect more similar patches than the cosine similarity; these results indicate that the Euclidean distance is more suitable to calculate the similarity score compared to the cosine similarity. To confirm the improvement of the modified CAM, moreover, we compared the quantitative performance of PEPSI with conventional and modified CAMs. As shown in Table~\ref{table:table9}, the modified CAM enhances the performance as compared to the conventional CAM, implying that the modified CAM is more appropriate to learn the relationship between background and hole regions. 

\section{Conclusion}
\label{sec6}
In this study, we have introduced a novel image inpainting model called PEPSI which overcomes the limitation of the two-stage coarse-to-fine network via the joint learning scheme. We provided qualitative and quantitative comparisons on CelebA-HQ and Place2 datasets. Experimental results revealed that PEPSI not only achieves superior performance as compared with conventional techniques, but also significantly reduces the computational time via a parallel decoding path and an effective joint learning scheme. Furthermore, we have introduced Diet-PEPSI which utilizes novel rate-adaptive convolutional layers to aggregate the global contextual information with low hardware costs. Experimental results shows that Diet-PEPSI preserves the performance of PEPSI while significantly reducing the hardware costs, which facilitates hardware implementation. Both networks are trained with the proposed RED and show visually plausible results in square holes as well as holes with an irregular shape. Therefore, it is expected that the proposed methods can be widely employed in various applications including image generation, style transfer, and image editing. 


\bibliographystyle{IEEEtran}
\bibliography{egbib.bib}

%

\begin{IEEEbiography}[{\includegraphics[width=1in,height=1.25in,clip,keepaspectratio]{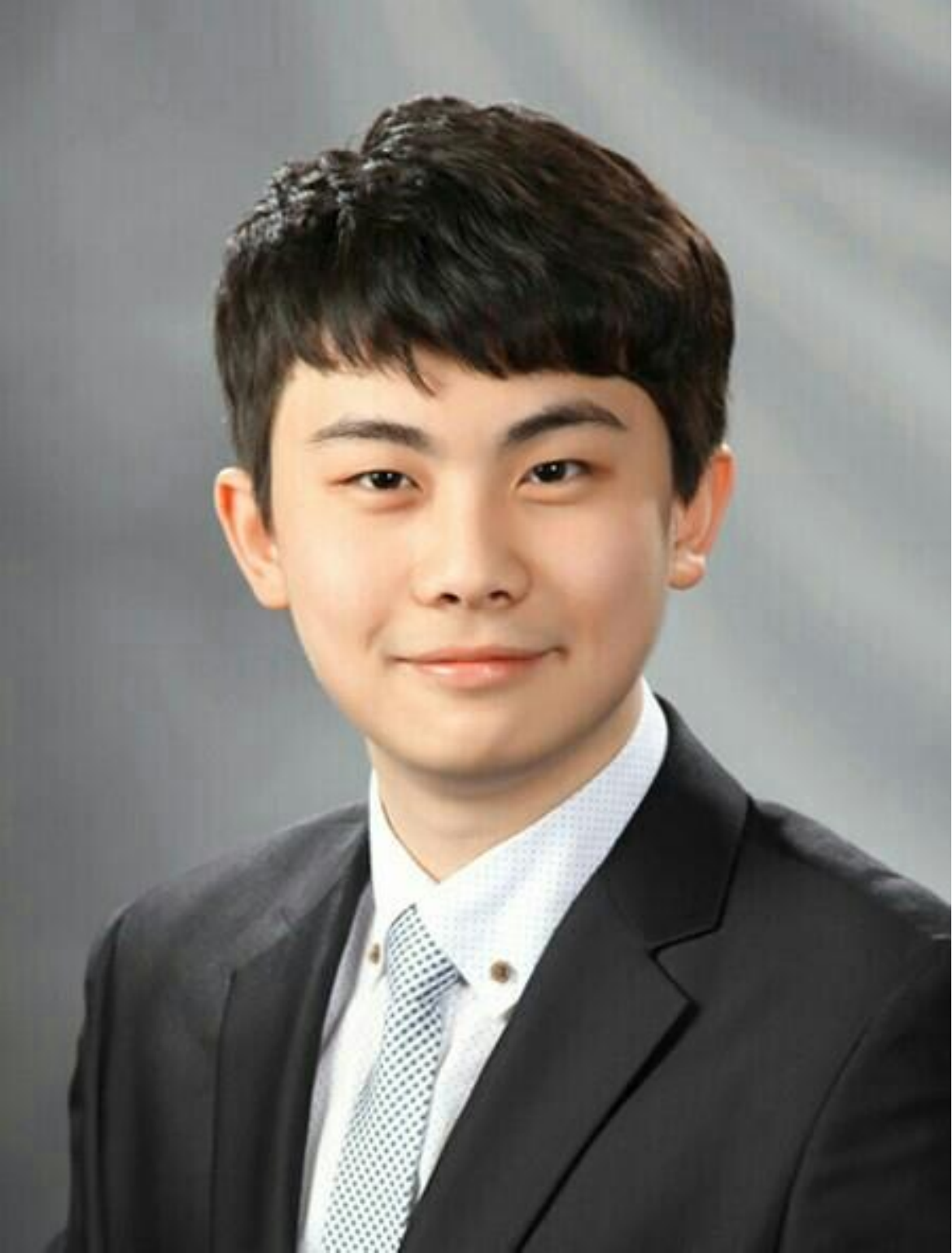}}]{Yong-Goo Shin}
received the B.S. and Ph.D. degrees in Electronics Engineering from Korea University, Seoul, Rep. of Korea, in 2014 and 2020, respectively. He is currently a research professor in the Department of Electrical Engineering of Korea University. His research interests are in the areas of digital image processing, computer vision, and artificial intelligence.
\end{IEEEbiography}

\begin{IEEEbiography}[{\includegraphics[width=1in,height=1.25in,clip,keepaspectratio]{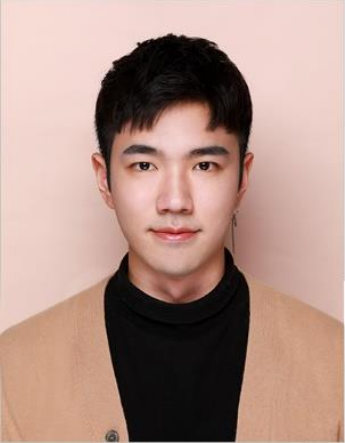}}]{Min-Cheol Sagong}
received his B.S. degree in Electrical Engineering from Korea University in 2018. He is currently pursuing his M.S. degree in Electrical Engineering at Korea University. His research interests are in the areas of digital signal processing, computer vision, and artificial intelligence.
\end{IEEEbiography}

\begin{IEEEbiography}[{\includegraphics[width=1in,height=1.25in,clip,keepaspectratio]{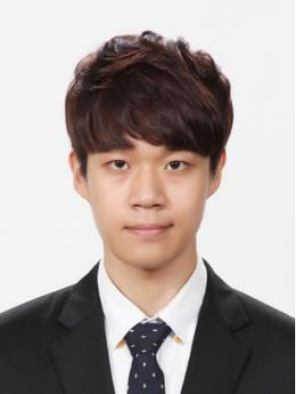}}]{Yoon-Jae Yeo}
received his B.S. degree in Electrical Engineering from Korea University in 2017. He is currently pursuing his Ph.D. degree in Electrical Engineering at Korea University. His research interests are in the areas of image processing, computer vision, and deep learning.
\end{IEEEbiography}

\begin{IEEEbiography}[{\includegraphics[width=1in,height=1.25in,clip,keepaspectratio]{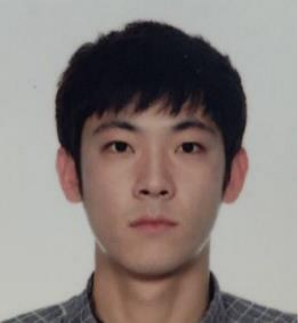}}]{Seung-Wook Kim}
received the B.S. and Ph.D. degrees in Electronics Engineering from Korea University, Seoul, Rep. of Korea, in 2012 and 2019, respectively. He is currently a research professor in the Department of Electrical Engineering of Korea University. His research interests are in the areas of image processing and computer vision based on deep learning.
\end{IEEEbiography}

\begin{IEEEbiography}[{\includegraphics[width=1in,height=1.25in,clip,keepaspectratio]{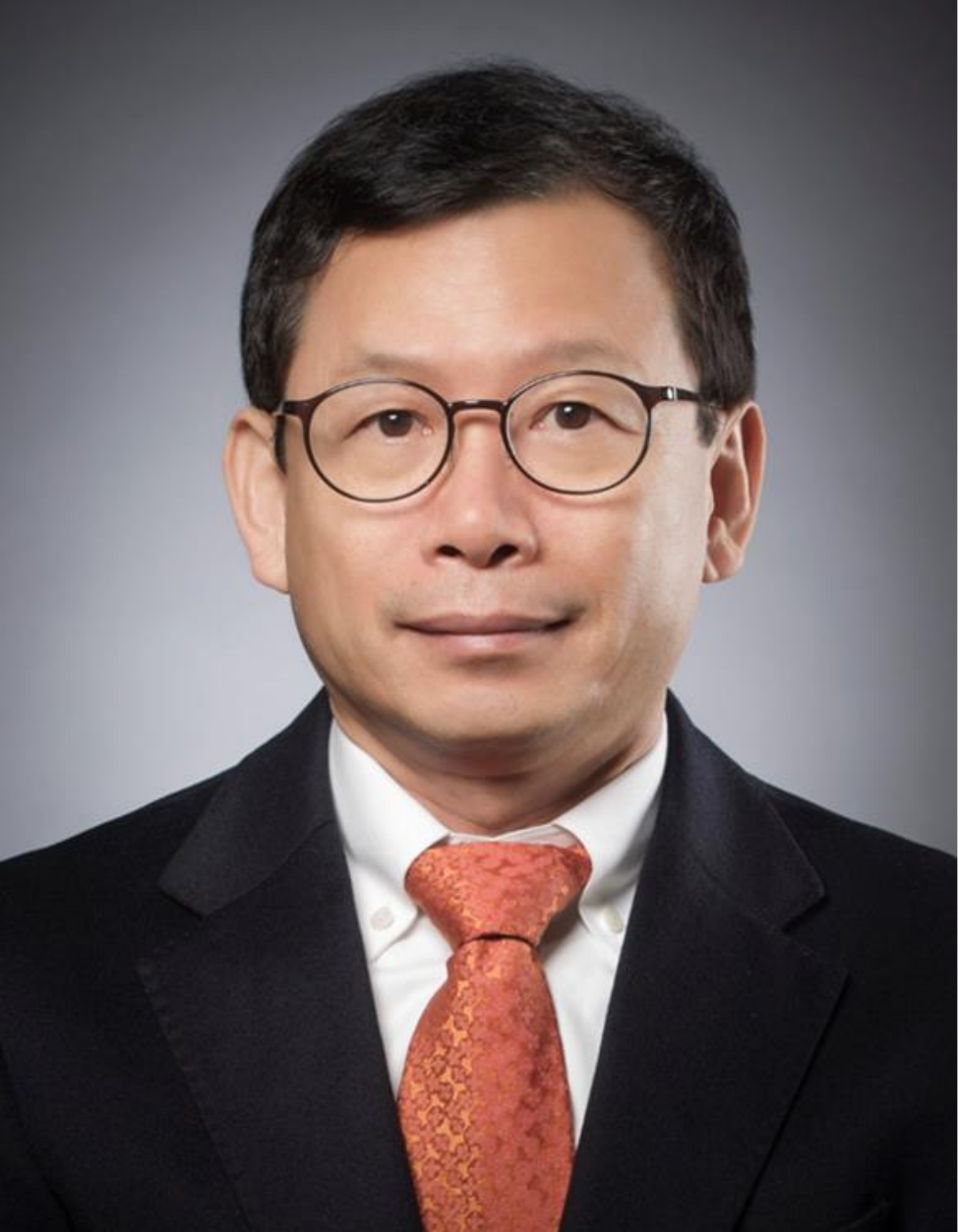}}]{Sung-Jea Ko}
(M’88-SM’97-F’12) received his Ph.D. degree in 1988 and his M.S. degree in 1986, both in Electrical and Computer Engineering, from State University of New York at Buffalo, and his B.S. degree in Electronic Engineering at Korea University in 1980. In 1992, he joined the Department of Electronic Engineering at Korea University where he is currently a Professor. From 1988 to 1992, he was an Assistant Professor in the Department of Electrical and Computer Engineering at the University of Michigan-Dearborn. He has published over 210 international journal articles. He also holds over 60 registered patents in fields such as video signal processing and computer vision. 

Prof. Ko received the best paper award from the IEEE Asia Pacific Conference on Circuits and Systems (1996), the LG Research Award (1999), and both the technical achievement award (2012) and the Chester Sall award from the IEEE Consumer Electronics Society (2017). He was the President of the IEIE in 2013 and the Vice-President of the IEEE CE Society from 2013 to 2016. He is a member of the National Academy of Engineering of Korea. He is a member of the editorial board of the IEEE Transactions on Consumer Electronics.
\end{IEEEbiography}

\end{document}